\documentclass[10pt,twocolumn,letterpaper]{article}

\usepackage{iccv}
\usepackage{times}
\usepackage{epsfig}
\usepackage{graphicx}
\usepackage{amsmath}
\usepackage{amssymb}

\usepackage{booktabs}
\usepackage{silence}
\usepackage{multirow}
\usepackage{enumitem}
\usepackage{subcaption}

\usepackage[pagebackref=true,breaklinks=true,colorlinks,bookmarks=false]{hyperref}

\iccvfinalcopy 

\def\iccvPaperID{5318} 

\ificcvfinal\fi

\begin{document}

\title{Interactive Image Segmentation with Cross-Modality Vision Transformers}

\author{Kun Li, George Vosselman, Michael Ying Yang\\
University of Twente}

\maketitle
\ificcvfinal\thispagestyle{empty}\fi

\begin{abstract}
   Interactive image segmentation aims to segment the target from the background with the manual guidance, which takes as input multimodal data such as images, clicks, scribbles, and bounding boxes. Recently, vision transformers have achieved a great success in several downstream visual tasks, and a few efforts have been made to bring this powerful architecture to interactive segmentation task. However, the previous works neglect the relations between two modalities and directly mock the way of processing purely visual information with self-attentions. In this paper, we propose a simple yet effective network for click-based interactive segmentation with cross-modality vision transformers. Cross-modality transformers exploits mutual information to better guide the learning process. The experiments on several benchmarks show that the proposed method achieves superior performance in comparison to the previous state-of-the-art models. The stability of our method in term of avoiding failure cases shows its potential to be a practical annotation tool. 
  The code and pretrained models will be released under \url{https://github.com/lik1996/iCMFormer}.
\end{abstract}

\section{Introduction}
\label{sec:intro}
Instance segmentation networks take a RGB-channel image as input and predict the segmentation mask in one single inference. Differently, interactive image segmentation is fed with not only the image but the interactions to identify the target of interest with sequential human-in-the-loops. This mechanism transforms interactive segmentation into a progressive coarse-to-fine dense prediction task, which has garnered significant interests of researchers working on related visual tasks such as image editing \cite{lee2020imageedit}, object selection \cite{ahmed2014objectselection}, medical image analysis \cite{shen2017medical}. Moreover, due to its class-agnostic predictions, interactive segmentation has the potential to serve as an annotation tool that generates large-scale labeled data for mask-level tasks such as semantic segmentation \cite{long2015fcn}, instance segmentation \cite{liu2018instanceseg} and autonomous driving \cite{siam2018autodrive}. Therefore, more and more efforts are put into this field from both academic and industrial communities.

Click-based interactive segmentation stands out by the advantage of simplicity and convenience. In the standard pipeline for interactive image segmentation, users first put a positive click on the target, and further add positive or negative clicks on the foreground or background, respectively, based on the current segmentation result. This iterative prediction process will not end until the segmentation meets the requirements.


\begin{figure}[t]
  \centering
  \includegraphics[width=1\linewidth]{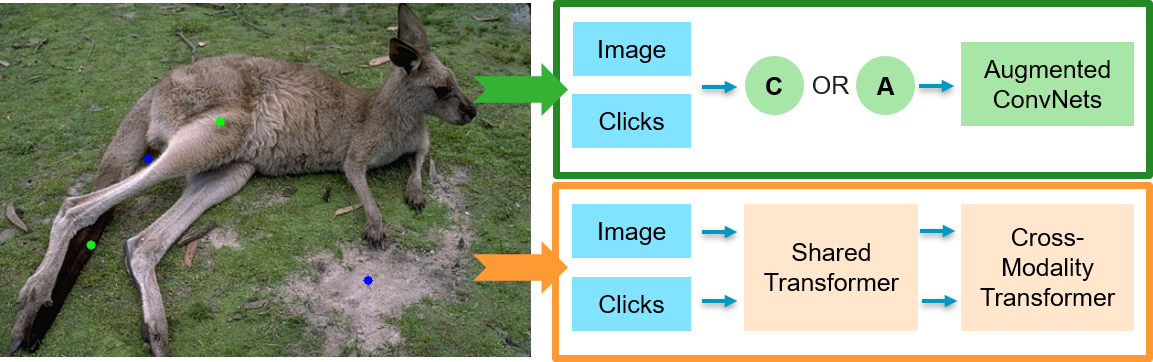}
  \caption{Illustration of our cross-modality transformers and the traditional incorporation in ConvNets. The green and blue dots denote the positive and negative clicks in the left part, respectively. The blue arrow represents one feeding path in the network. The green box shows the simple combination strategies (e.g., concatenation, addition) adopted in the previous models. Our method considers the cross-modality guidance with different transformer blocks, as shown in the orange box.}
  \label{fig:illustration}
  \vspace{-5mm}
\end{figure}

Over the last few years, click-based interactive segmentation has made great strides in various directions such as sampling strategy \cite{xu2016dios}, click encoding \cite{majumder2019content}, powerful backbones \cite{chen2021cdnet}, local refinements \cite{lin2022focuscut, wei2023fcf}, and computational optimization \cite{chen2022focalclick}. The green box of Fig.~\ref{fig:illustration} shows the architecture of most existing methods. The positive and negative clicks are represented as 2D masks by the same size as the input image. To make use of the pretrained models for robust feature extraction, these methods augment the weights of certain layers for the concatenated or element-size summarized image and click masks \cite{sofiiuk2022ritm}. However, they utilize two-modality input indiscriminately with purely visual information processing. In practical, the discrete clicks (either distance maps or disk maps) should be seen as a guidance signal in the process of image segmentation. Meanwhile, the value ranges between images and click masks do not match well if directly concatenating or adding them together in the early stages. Based on the above concerns in mind, a better incorporation method between image and clicks is in high demand for interactive segmentation. 


In this paper, we propose \textit{interactive Cross-Modality TransFormer} (iCMFormer), a vision transformer based method with cross-modality attentions between image and clicks (shown in the orange box of Fig.~\ref{fig:illustration}). To alleviate the mismatching problem in the early stage, we use a parallel structure for both modalities with shared vision transformer blocks. We propose cross-modality transformers to extract guidance signals, which can help improve focus on the target locations. By incorporating another group of vision transformers for high-level semantic information extraction, the fused features from both branches can be finely tuned before going through the segmentation head. Inspired by the progressive downsampling operations in ConvNets \cite{he2016resnet,lin2020fca} for larger receptive fields, hierarchical vision transformers address multi-scale problem with similar stages. Our proposed cross-modality transformers are flexible to be added into the hierarchical structure such as Swin-Transformer \cite{liu2021swin} to improve the results. We evaluate our method on four datasets through a series of experiments, and the results show the superior performance of iCMFormer compared with the existing methods. Our \textbf{main contributions} of this paper are summarized as follows:
\begin{itemize}
    \item Our iCMFormer is the first network that takes the modality issue into account with vision transformers for interactive segmentation. The proposed simple yet effective cross-modality transformers utilize the guidance information to generate robust results.
    \item The proposed cross-modality transformers are flexibly integrated into a hierarchical architecture to address the multi-scale problem.
    \item Our method achieves the state-of-the-art performance on four benchmarks, which can be explored as a practical annotation tool for other visual tasks.
\end{itemize}

\section{Related Work}
\label{related work}

\noindent
\textbf{Interactive Segmentation Methods.} Interactive segmentation (IS) is a quite active research field, which involves progressive interactions between humans and machines. Early works \cite{grady2006random,li2004lazy,rother2004grabcut} address this problem from the perspective of optimization. However, these works fail to handle complex surroundings by only relying on the low-level features. Since ConvNets show their power on extracting robust features from images, some IS methods adopt the successful backbones \cite{he2016resnet,long2015fcn,simonyan2014vgg} to improve the segmentation results. DIOS \cite{xu2016dios} is the first work to bring deep learning techniques to IS, and proposes a classical sampling strategy to simulate positive and negative clicks for training. Not restricted in clicks, more interaction formats (e.g., scribbles \cite{asad2022scrribles}, polygons \cite{acuna2018poligonrnn+}, bounding boxes \cite{xu2017deepgrabcut}) have also been explored. DEXTR \cite{maninis2018dextr} makes use of four extreme points: the left-most, right-most, top-most, and bottom-most pixels to specify the target from the background. ITIS \cite{mahadevan2018itis} proposes a new online iterative sampling strategy based on the regions from the current incorrect predictions, which has been improved in RITM \cite{sofiiuk2022ritm} with less computational resources. Not only the global segmentation, but further refinements are beneficial to obtain high-quality results. Backpropagating refinement scheme \cite{jang2019brs,sofiiuk2020fbrs} minimizes a discrepancy between the input map and predicted mask for optimization. FocalClick \cite{chen2022focalclick}, FocusCut \cite{lin2022focuscut} and FCF \cite{wei2023fcf} try to modify the segmentation results from the local perspective. From other aspects for IS, EMC \cite{du2023emc} reduces the computational cost via a lightweight mask correction network. GPCIS \cite{zhou2023gaussion} formulates IS as a Gaussian process classification model on each image. However, these methods neglect the modality issue but attempt to improve the results through complex attention modules or local refinements. Differently, we explore simple vision transformer backbones equipped with cross-modality transformers for IS.


\noindent
\textbf{Vision Transformers.} Attention-based transformers \cite{vaswani2017transformer} have achieved great performance in the field of natural language processing (NLP), which has attracted lots of interests in computer vision community. The original ViT \cite{dosovitskiy2020vit} brings the self-attention transformers to image classification task with sequentially processing for smaller image patches. However, the plain transformers with encoder-decoder architecture are insufficient for the dense prediction tasks such as semantic segmentation. Various hierarchical vision transformers \cite{chu2021twins,wang2021pyramid-t,xie2021segformer,yuan2021hrformer} have been proposed to solve the problem with different network designs. These methods are inspired by the ideas from successful ConvNets such as hierarchical structure, multi-scale and multi-path designs, pooling and down-sampling operations. For instance, Swin-Transformer \cite{liu2021swin} handles the reduced resolution feature maps with high-level semantic information, and captures multi-stage features to obtain good results. Correspondingly, the hierarchical structure can be used with our proposed cross-modality transformers to address the multi-scale problem.

\noindent
\textbf{Multimodal Learning.} 
In the last decade, we have witnessed the rising and fast pace developments of deep learning models for multimodal streams such as vision\&text\cite{antol2015vqa,zhou2020unifiedvqa}, video\&audio \cite{ngiam2011multimodal} and RGB\&Lidar \cite{qian2020lidar}. Normally, these tasks need a shared representation approach between modalities, as well as the cross-modality learning for the feature fusion. The previous interactive segmentation methods \cite{lin2020fca,majumder2019content,xu2016dios}
only take the interactions as another format of image mask (e.g., binary disk map, Gaussian map, or distance map) and seldom study the relations between modalities. In our work, the multimodal information is learnt with the proposed cross-attention transformers.
\begin{figure*}[ht]
  \centering
  \includegraphics[width=0.85\linewidth]{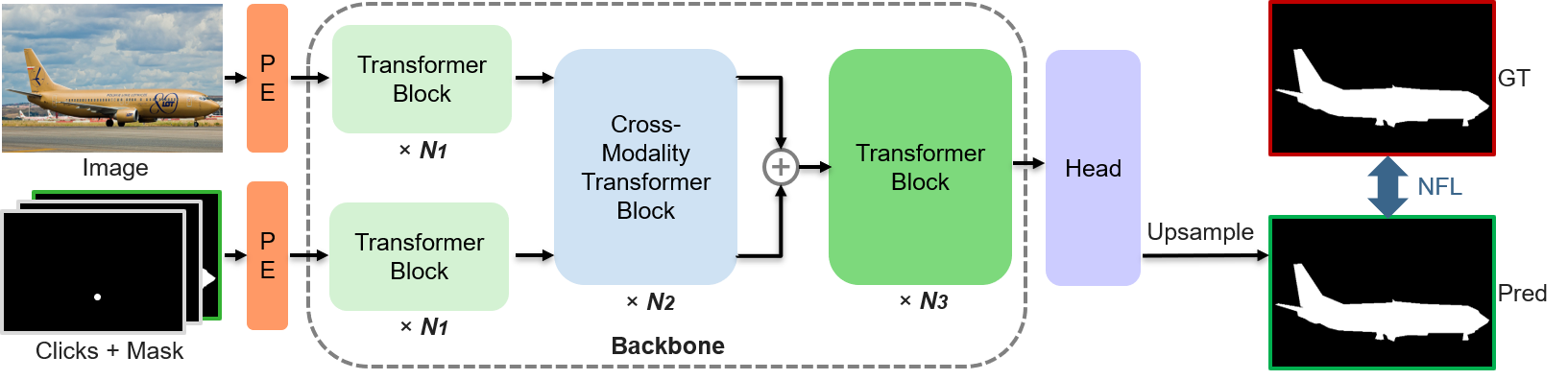}
  \vspace{-2mm}
  \caption{The overall architecture of our method. The positive and negative clicks (transformed into two-channel disk maps) plus the previous segmentation mask are concatenated as input for the interaction branch. PE and NFL denote the patch embedding operation and normalized focal loss, respectively. For brevity, the positional embedding is not shown here. We provide two backbones with the similar pipeline (see Sec.~\ref{backbone} for details). The light green part shows the shared self-attention transformer group for two branches (6 blocks for ViT-B and 2 plus 2 blocks for Swin-B), while the dark green part shows the second transformer group for the combined input (6 blocks for ViT-B and 18 plus 2 blocks for Swin-B). The number of cross-modality transformers in the light blue part is set to 3 and 4 for ViT-B and Swin-B, respectively. The segmentation head coupled with upsampling operations processes the attended features to obtain the final prediction.}
  \label{fig:pipeline}
  \vspace{-2mm}
\end{figure*}

\section{Method}
\label{method}
We propose an interactive image segmentation method on the basis of vision transformers. In this section, we first introduce the network with plain and hierarchical structure, respectively. Then we elaborate the cross-modality attentions for learning relationships between images and clicks. Finally, we explain the iterative training scheme and details about click simulations.
\subsection{Effective Network}
\label{backbone}
The architecture of the proposed network for interactive segmentation is shown in Fig.~\ref{fig:pipeline}. We retain the original blocks and corresponding hyper-parameters for both plain and hierarchical transformers. Instead, we add the cross-modality attention blocks (introduced in Sec.~\ref{cross-modality}) in the middle stage of these transformers. On the basis of the backbones, a segmentation head is adopted to obtain dense predictions. More details can be found in the supplementary material.

\vspace{2mm}
\noindent
\textbf{Backbones.} To extract the features from images and clicks, we employ two powerful vision transformers as our backbones: plain vision transformer \cite{dosovitskiy2020vit} and Swin Transformer \cite{liu2021swin}. Plain vision transformer (ViT) is a classical self-attention network by splitting the images into smaller patches with positional embeddings, which is inspired by the original transformer \cite{vaswani2017transformer} for sequential text processing. Then these patches are further flattened and projected into a linear space as a vector that serves as the input for transformers. We divide the 12 transformer blocks from the base version of vision transformer (ViT-B) into 2 groups, and add 3 blocks of the proposed cross-modality transformer between them. The other backbone is Swin-Transformer, which has a hierarchical architecture with linear computational complexity through window and shifted-window self-attentions. Similarly, we divide the base Swin-Transformer (Swin-B) into 2 groups and add 4 proposed blocks. Note that the first 2 stages (2 plus 2 transformers) of Swin-B are grouped while the others (18 plus 2 transformers) as the second group. For both ViT-B and Swin-B backbones, the input is fed into a shared network consisting of the first group of transformers, which processes the data for different branches, including images and clicks. After the followed cross-modality transformers, the image features and click features are combined with an element-wise addition, as the input for the next group of transformers. About the click encoding for networks, RITM \cite{sofiiuk2022ritm} has concluded that the disk maps perform better than others (distance maps and Gaussian maps). We directly employ the disk maps (radius equals 5) in our work.

\vspace{2mm}
\noindent
\textbf{Segmentation Head.} As the hierarchical transformer Swin-B has a large receptive field, it is unnecessary to design complex hand-crafted components like original segmentation follow-ups. We employ the simple segmentation head from Segformer \cite{xie2021segformer} in our work. Specifically, it consists of 4 MLP steps: unification on the channel dimension for the multi-scale features from the backbones, upsampling the features to the same resolution, fusion based on the concatenated features, and prediction with a sigmoid for the final segmentation result. To unify the framework for different backbones, we add 4 convolution layers (inspired by ViTDet \cite{li2022vitdet}) for the last output from the ViT-B, and adopt the same segmentation head. After upsamlping operations to obtain the same resolution of the original image, the probability map for the foreground prediction is generated. 
\begin{figure}[t!]
  \centering
  \begin{subfigure}[t]{0.4\linewidth}
  \includegraphics[width=0.8\linewidth]{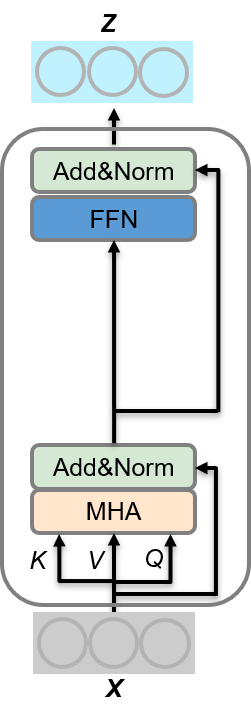}
  \subcaption{Self-attention.}
  \label{fig:self-a}
  \end{subfigure}
  \begin{subfigure}[t]{0.45\linewidth}
  \includegraphics[width=1\linewidth]{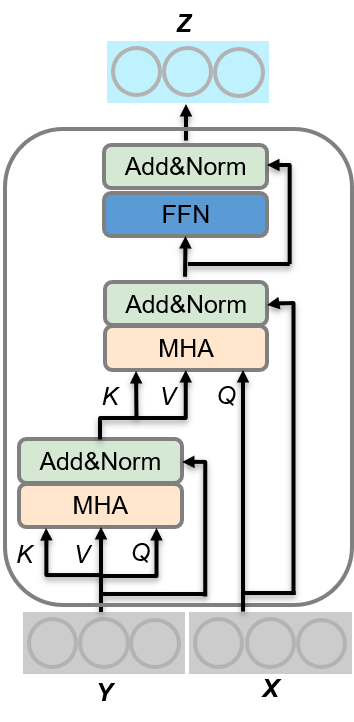}
  \subcaption{Cross-attention.}
  \label{fig:cross-a}
  \end{subfigure}
  \caption{Self-attention only takes one modality input while cross-attention takes input from both image and clicks. $Z$, $X$ and $Y$ denote attended outputs, image and click features, respectively.}
  \vspace{-3mm}
  \label{fig:cross}
\end{figure}

\subsection{Cross-Modality Attention}
\label{cross-modality}
Multi-head attention (MHA) is the basic function in the original transformer blocks, which takes in the query, key, and value to capture different focuses. The function outputs the summation over the values with weighted attentions obtained from the scaled dot products between queries and keys. Note that the $Q,K,V$ indicating the queries, keys, and values, respectively, are obtained from the same input features (shown in Fig.~\ref{fig:self-a}, which is also called self-attention \cite{vaswani2017transformer}). Take one head as an example for the self-attention:
\begin{equation}
    f_{self} = A(Q,K,V) = {\rm Softmax}(\frac{QK^T}{\sqrt{d}})V,
\end{equation}
where $d$ represents the dimension of keys and values.

Inspired by some vision-language works \cite{yu2019mcan,yuan2022cross-ic}, we propose a cross-modality transformer block (see in Fig.~\ref{fig:cross-a}) for interactive segmentation. A cross-modality block takes two groups of features $X$ and $Y$ from images and clicks, where one modality $Y$ guides the learning for the other one $X$. Specifically, the block consists of 2 steps of multi-head attentions (MHA): self-attentions on the $Q,K,V$ from $Y$, and the cross-attentions on the $Q$ (from $X$) with $K$ and $V$ (both from $Y$), where it learns to capture the cross-modal relationships. The cross-attention is given by:
\begin{equation}
    f_{cross} = A(Q_x,K_y,V_y) = {\rm Softmax}(\frac{Q_xK_y^T}{\sqrt{d_y}})V_y,
\end{equation}
where $Q_x$ represents the queries from $X$ while $K_y$, $V_y$ and $d_y$ denote the keys, values and dimension of keys from $Y$. Then it follows a feed-forward network (FFN) with ReLU activation and Dropout like a standard transformer block.  
\subsection{Iterative Training Scheme}
Before introducing the training scheme for deep interactive segmentation networks, we take a deep dive into the interactions involved in a human-in-the-loop mechanism. Normally, the first click (always positive one) should be put into the centre of the target while every new click is placed in the regions where the model has made incorrect predictions. Whether a new click is positive or negative is decided by humans based on the analysis on the current segmentation result. Therefore, interactive segmentation is a progressive refinement method based on a set of sequential clicks.

However, previous methods \cite{li2018latent,lin2020fca,majumder2019content} ignore the sequential information by adopting random sampling strategy \cite{xu2016dios} in the training stage. RITM \cite{sofiiuk2022ritm} propose a novel iterative sampling strategy, which generates the next click in the cluster centre of the largest incorrect prediction region after morphological erosion operation. To reduce computation in the training, the maximum number of iterative clicks is set to 3. We employ the similar click simulation strategy with RITM, and make a small change on the selection of iterative click's position. Specifically, we combine the centre point and random point near the borders of the mislabeled regions to fit humans' behaviors better.

In addition, we incorporate the segmentation mask from the last iterative step as an additional channel for the click branch, which has been proved as prior information \cite{mahadevan2018itis} to improve the results. Note that we feed an empty mask for the first iteration. We also take the Normalized Focal Loss \cite{sofiiuk2019nfl} (NFL) as the loss function for the training following recent works \cite{chen2022focalclick,sofiiuk2022ritm}, which converges faster and more robustly.

\begin{table*}
  \caption{Evaluation results on GrabCut \cite{rother2004grabcut}, Berkeley \cite{martin2001berkely}, SBD \cite{hariharan2011sbd} and DAVIS \cite{perazzi2016davis} datasets. NoC85 and NoC90 denote the average numbers of clicks to reach a target IoU. The best results are \textbf{bold} while the second best are \underline{underlined}. Note that $\S$, $\dagger$ and $\ddagger$ represent the models trained on PASCAL \cite{everingham2009pascal}, SBD, and COCO \cite{lin2014coco} + LVIS \cite{gupta2019lvis}, respectively.}
  \vspace{-2mm}
  \centering
  \setlength\tabcolsep{4pt}
  \begin{tabular}{p{2.5cm}cccccccccc}
  \toprule
  \multirow{2}*{Method} & \multirow{2}*{Year} & \multirow{2}*{Backbone} & \multicolumn{2}{c}{GrabCut} & \multicolumn{2}{c}{Berkeley} & \multicolumn{2}{c}{SBD} & \multicolumn{2}{c}{DAVIS}\\
 & & & NoC85 &  NoC90 & NoC85 &  NoC90 & NoC85 &  NoC90 & NoC85 &  NoC90\\
 \midrule
  DIOS\cite{xu2016dios}$\S$   & CVPR16 & FCN & - & 6.04 & - & 8.65 & - & - & - & 12.58 \\
  RIS-Net\cite{liew2017ris-net}$\S$   & ICCV17 & FCN & - & 5.00 & - & - & 6.03 & - & - & - \\
  FCA-Net\cite{lin2020fca}$\S$   & CVPR20 & ResNet-101 & - & 2.08 & - & 3.92 & - & - & - & 7.57 \\
  \midrule\hline
  LD\cite{li2018latent}$\dagger$   & CVPR18 & VGG-19 & 3.20 & 4.79 & - & - & 7.41 & 10.78 & 5.05 & 9.57 \\
  BRS\cite{jang2019brs}$\dagger$  & CVPR19 & DenseNet & 2.60 & 3.60 & - & 5.08 & 6.59 & 9.78 & 5.58 & 8.24\\
  f-BRS\cite{sofiiuk2020fbrs}$\dagger$   & CVPR20 & ResNet-101 & 2.30 & 2.72 & - & 4.57 & 4.81 & 7.73 & 5.04 & 7.41\\
  CDNet\cite{chen2021cdnet}$\dagger$   & ICCV21 & ResNet-50 & 2.22 & 2.64 & - & 3.69 & 4.37 & 7.87 & 5.17 & 6.66\\
  RITM\cite{sofiiuk2022ritm}$\dagger$   & ICIP22 & HRNet-18 & 1.76 & 2.04 & - & 3.22 & 3.39 & 5.43 & 4.94 & 6.71\\
  FocalClick\cite{chen2022focalclick}$\dagger$   & CVPR22 & HRNet-18s-S2 & 1.86 & 2.06 & - & 3.14 & 4.30 & 6.52 & 4.92 & 6.48\\
  FocalClick\cite{chen2022focalclick}$\dagger$   & CVPR22 & SegF-B0-S2 & 1.66 & 1.90 & - & 3.14 & 4.34 & 6.51 & 5.02 & 7.06\\
  FocusCut\cite{lin2022focuscut}$\dagger$ & CVPR22 & ResNet-101 & \underline{1.46} & 1.64 & - & 3.01 & 3.40 & \underline{5.31} & 4.85 & 6.22 \\
PseudoClick\cite{liu2022pseudoclick}$\dagger$ & ECCV22 & HRNet-18 & - & 2.04 & - & 3.23 & - & 5.40 & 4.81 & 6.57 \\
  GPCIS\cite{zhou2023gaussion}$\dagger$ & CVPR23 & HRNet-18s-S2 & 1.74 & 1.94 & 1.83 & 2.65 & 4.28 & 6.25 & 4.62 & 6.16 \\
  GPCIS\cite{zhou2023gaussion}$\dagger$ & CVPR23 & SegF-B0-S2 & 1.60 & 1.76 & 1.84 & 2.70 & 4.16 & 6.28 & 4.45 & 6.04 \\
  EMC\cite{du2023emc}$\dagger$ & CVPR23 & HRNet-18 & 1.74 & 1.84 & - & 3.03 & 3.38 & 5.51 & 5.05 & 6.71 \\
  FCF\cite{wei2023fcf}$\dagger$ & CVPR23 & ResNet-101 & 1.64 & 1.80 & - & 2.84 & \underline{3.26} & 5.35 & 4.75 & 6.48 \\
  Ours$\dagger$ & 2023 & ViT-B & \textbf{1.36} & \textbf{1.42} & \textbf{1.42} & \underline{2.52} & 3.33 & \underline{5.31} & \textbf{4.05} & \underline{5.58}\\
  Ours$\dagger$ & 2023 & Swin-B & \underline{1.46} & \underline{1.50} & \underline{1.52} & \textbf{2.32} & \textbf{3.21} & \textbf{5.16} & \underline{4.25} & \textbf{5.55}\\
  \midrule\hline
  RITM\cite{sofiiuk2022ritm}$\ddagger$   & ICIP22 & HRNet-18 & 1.42 & 1.54 & - & 2.26 & 3.80 & 6.06 & 4.36 & 5.74\\
  RITM\cite{sofiiuk2022ritm}$\ddagger$   & ICIP22 & HRNet-32 & 1.46 & 1.56 & - & 2.10 & 3.59 & 5.71 & 4.11 & 5.34\\
  FocalClick\cite{chen2022focalclick}$\ddagger$   & CVPR22 & HRNet-32-S2 & 1.64 & 1.80 & - & 2.36 & 4.24 & 6.51 & 4.01 & 5.39\\
  FocalClick\cite{chen2022focalclick}$\ddagger$   & CVPR22 & SegF-B0-S2 & \underline{1.40} & 1.66 & - & 2.27 & 4.56 & 6.86 & 4.04 & 5.49\\ 
 PseudoClick\cite{liu2022pseudoclick}$\ddagger$ & ECCV22 & HRNet-32 & - & 1.50 & - & 2.08 & - & 5.54 & 3.79 & 5.11 \\
  EMC\cite{du2023emc}$\ddagger$ & CVPR23 & SegF-B3 & 1.42 & \underline{1.48} & - & 2.35 & 3.44 & 5.57 & 4.49 & 5.69 \\
  FCF\cite{wei2023fcf}$\ddagger$ & CVPR23 & HRNet-18 & \textbf{1.38} & \textbf{1.46} & - & 1.96 & 3.63 & 5.83 & 3.97 & 5.16 \\
  
  Ours$\ddagger$ & 2023 & ViT-B & 1.42 & 1.52 & \textbf{1.40} & \textbf{1.86} & \underline{3.29} & \underline{5.30} & \textbf{3.40} & \underline{5.06}\\
  Ours$\ddagger$ & 2023 & Swin-S & 1.46 & 1.60 & 1.49 & \underline{1.93} & 3.34 & 5.35 & \underline{3.46} & 5.07 \\
  Ours$\ddagger$ & 2023 & Swin-B & 1.42 & 1.54 & \underline{1.42} & 2.03 & \textbf{3.12} & \textbf{5.11} & 3.48 & \textbf{5.03}\\
  \bottomrule
  \end{tabular}
  \label{tab:comparison with SOTA}
  \vspace{-2mm}
\end{table*}
\section{Experiments}

\subsection{Experiment Setup}
\label{setup}
\noindent
\textbf{Datasets.} We evaluate our proposed interactive segmentation method on four widely used datasets, and employ one combination dataset for large-scale training:

\begin{itemize}[leftmargin=*]
    \item \textbf{GrabCut \cite{rother2004grabcut}.} The dataset contains 50 images and provides one single instance mask for each image.
    \item \textbf{Berkeley \cite{martin2001berkely}.} The dataset provides 96 images and 100 instance masks, and some objects are hard to distinguished from the similar background. 
    \item \textbf{SBD \cite{hariharan2011sbd}.} The dataset is divided into two subsets for object segmentation task (training: 8498 images and 20172 instances, validation: 2857 images and 6671 instances). We train the models on the training set and evaluate the performances on the validation set like others \cite{chen2022focalclick,jang2019brs,sofiiuk2022ritm}.
    \item \textbf{DAVIS \cite{perazzi2016davis}.} The dataset is designed for video semantic segmentation. We take the same 345 frames from the labeled 50 videos for evaluation like \cite{jang2019brs}. 
    \item \textbf{COCO\cite{lin2014coco} + LVIS \cite{gupta2019lvis}.} Following \cite{sofiiuk2022ritm}, we take the combined version of COCO and LVIS with higher annotation quality for large-scale training, which contains 118K images with 1.2M instances.
\end{itemize}

\begin{figure*}[ht]
  \centering
    \includegraphics[trim= 10pt 0pt 10pt 14pt, clip=True, width=0.47\linewidth]{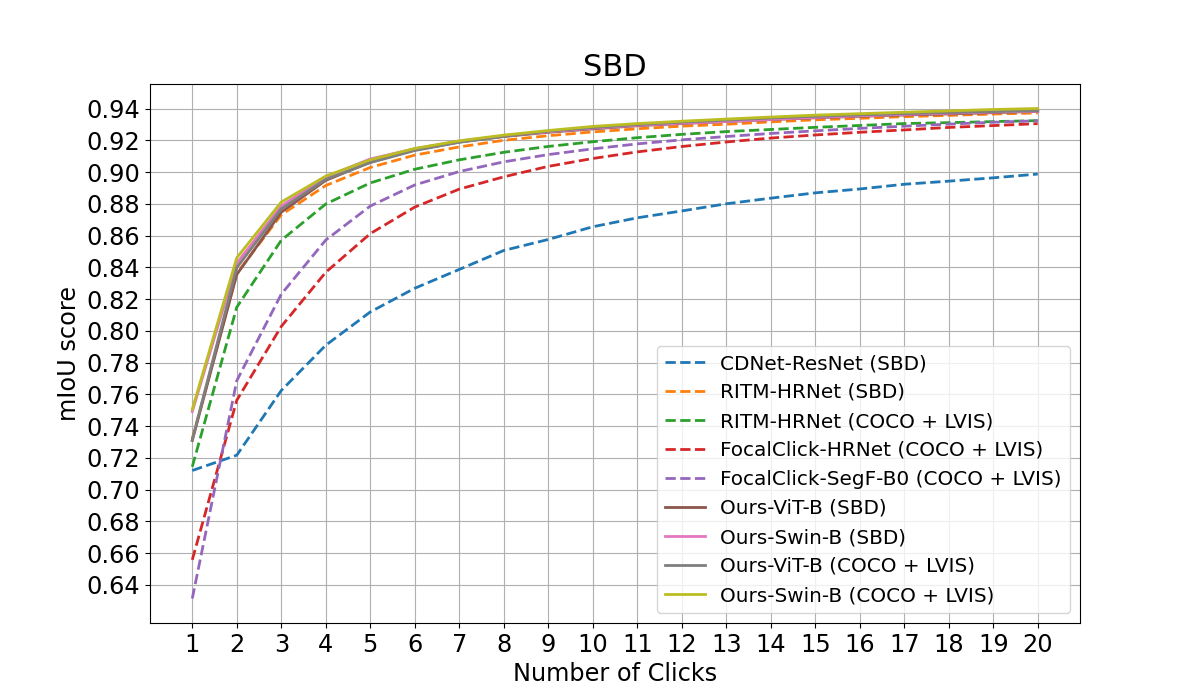}
  \hspace{2mm}
    \includegraphics[trim= 10pt 0pt 10pt 14pt, clip=True, width=0.47\linewidth]{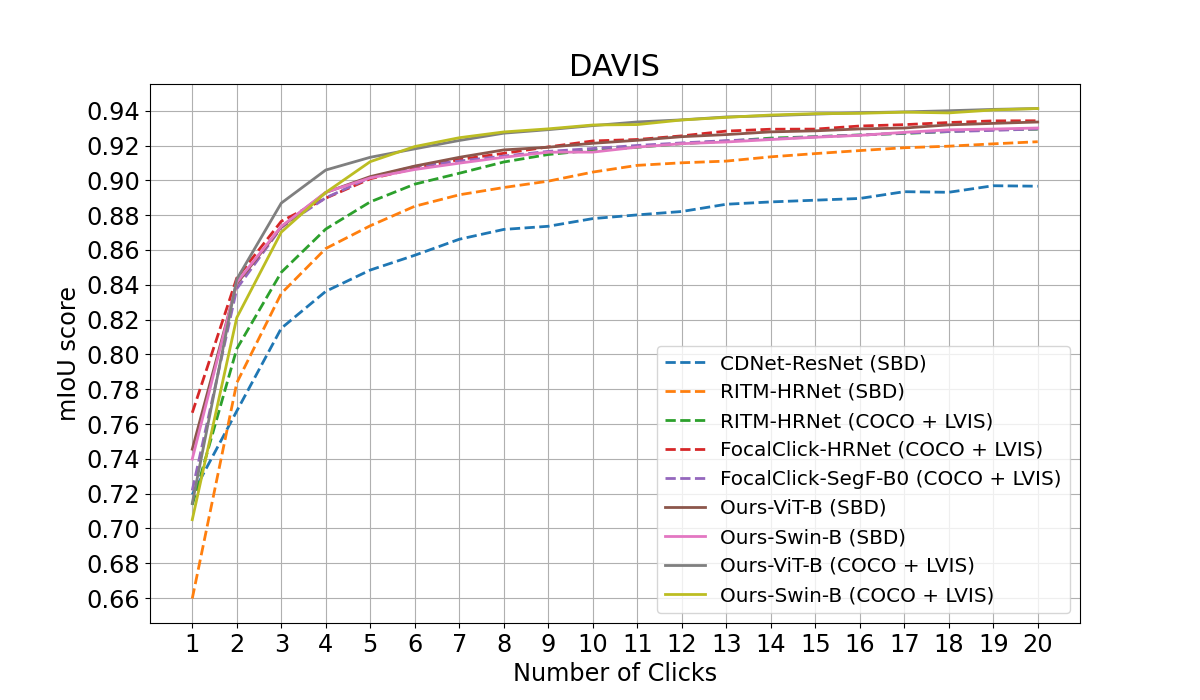}
  \vspace{-3mm}
  \caption{Convergence analysis of mean IoU curves for varying number of clicks. The evaluation results on SBD \cite{hariharan2011sbd} and DAVIS \cite{perazzi2016davis} are provided. The higher starting point typically leads to better results with the first positive click. A steeper slope indicates that the method requires fewer clicks to achieve better segmentation results.}
   \vspace{-4mm}
  \label{fig:anaysis}
\end{figure*}

\noindent
\textbf{Evaluation Protocol.} To evaluate the proposed method, two kinds of inference ways are employed in this paper: manual evaluation to qualitatively access the real interactive segmentation results and automatic evaluation based on the simulated clicks to make a quantitative comparison with the others. As for the automatic evaluation, the first click (compulsively positive one to indicate the target) is sampled in the centre of the target object, while the next click is always selected from the largest error region by comparing the current prediction mask with the ground truth.

For the metrics, mean Intersection over Union (mIoU) is adopted in our work as a common image segmentation evaluation metric. In addition, Number of Clicks (NoC) is used to evaluate the interaction efforts for reaching a certain IoU threshold within the maximum clicks. Number of Failures (NoF) means the number of instances that the model fails to obtain a corresponding IoU after the maximum round of clicks, which reflects the stability of the method. We set two IoU thresholds (85\% and 90\%) and 20 clicks as the upper bound for interactions, which are consistent with the previous works \cite{chen2022focalclick,li2018latent,lin2022focuscut,lin2020fca,xu2016dios}.

\noindent
\textbf{Implementation Details.} All the experiments are implemented on the PyTorch platform with 2 A40 GPUs. For different transformer backbones including ViT \cite{dosovitskiy2020vit} and Swin \cite{liu2021swin}, we use the pretrained models from the official repositories. During training, we employ several data augmentation strategies: random flipping, rotation, cropping as well as random resizing with the scale from 0.75 to 1.25. We apply Adam optimizer with $\beta_1=0.9$ and $\beta_2=0.99$. Our models are trained on SBD \cite{hariharan2011sbd} and COCO \cite{lin2014coco} + LVIS \cite{gupta2019lvis} with 55 and 85 epochs, respectively. We set batch size to 24, the initial learning rate as 0.00005 and decrease it 10 times after the epoch of 50.
\subsection{Comparison with State-of-the-Art}
We compare our results on four benchmarks with previous click-based interactive segmentation methods in terms of the mentioned evaluation metrics. Note that the maximum number of clicks is set as 20 for NoC@85 and NoC@90 even when the results cannot reach the target IoU, which is consistent with the other works \cite{chen2021cdnet,lin2020fca,xu2016dios}.

\begin{table}
  \caption{Comparison with previous models trained on SBD \cite{hariharan2011sbd} in term of number of failures (NoF) that cannot reach the target IoU after 20 clicks, denoted as $\geq$20@90.}
  \centering
  \setlength\tabcolsep{5pt}
  \begin{tabular}{p{2cm}ccc}
  \toprule
  \multirow{2}*{Method} & \multicolumn{1}{c}{Berkeley} & \multicolumn{1}{c}{SBD} & \multicolumn{1}{c}{DAVIS}\\
 & $\geq$20@90 & $\geq$20@90 & $\geq$20@90\\
  \midrule
  BRS\cite{jang2019brs}  & 10 & - & 77\\
  f-BRS\cite{sofiiuk2020fbrs}  & 2 & 1466 & 78\\
  CDNet\cite{chen2021cdnet}  & - & - & 65\\
  FocusCut\cite{lin2022focuscut} & - & - & 57 \\
  FCF\cite{wei2023fcf} & 3 & - & 59\\
  Ours-ViT-B & 2 & \textbf{693} & \textbf{53}\\
  Ours-Swin-B & \textbf{1} & 698 & \textbf{53}\\
  \bottomrule
  \end{tabular}
  \label{tab:comparison with SOTA for nums larger than 20}
  \vspace{-3mm}
\end{table}

\begin{table}
  \caption{Computation comparison with different models in terms of parameters, FLOPs and inference speed. The inference speed is evaluated by average time per click on GrabCut \cite{rother2004grabcut}. Note that as the input image size will influence the numbers, we report the sizes as well.}
  \centering
  \setlength\tabcolsep{4pt}
  \begin{tabular}{p{2.5cm}cccc}
    \toprule
    Model & Size & \# Params & \# FLOPs & SPC\\
    \midrule
    ResNet-101\cite{lin2022focuscut} & 384 & 59.35M & 102.02G & 384ms\\
    HRNet-18s\cite{sofiiuk2022ritm} & 400 & 4.22M & 17.84G & 64ms\\
    HRNet-18\cite{sofiiuk2022ritm} & 400 & 10.03M & 30.80G & 70ms\\
    HRNet-32\cite{sofiiuk2022ritm} & 400 & 30.95M & 82.84G & 84ms\\
    SegF-B0-S2\cite{chen2022focalclick} & 256 & 3.72M & 3.54G & 42ms\\
    SegF-B3-S2\cite{chen2022focalclick} & 256 & 45.66M & 25.34G & 76ms\\
    Ours-ViT-B & 448 & 124.81M & 297.54G & 78ms\\
    Ours-Swin-S & 224 & 68.14M & 106.74G & 74ms\\
    Ours-Swin-B & 384 & 104.25M & 153.78G & 86ms\\
    \bottomrule
  \end{tabular}
  \label{tab:computation comparison}
  \vspace{-4mm}
\end{table}

\begin{figure*}[ht]
  \centering
  \begin{subfigure}[t]{0.22\linewidth}
    \includegraphics[width=1\linewidth]{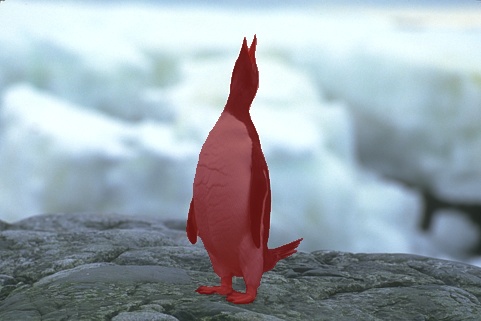}
    \subcaption*{GT}
  \end{subfigure}
  \hspace{2mm}
  \begin{subfigure}[t]{0.22\linewidth}
    \includegraphics[width=1\linewidth]{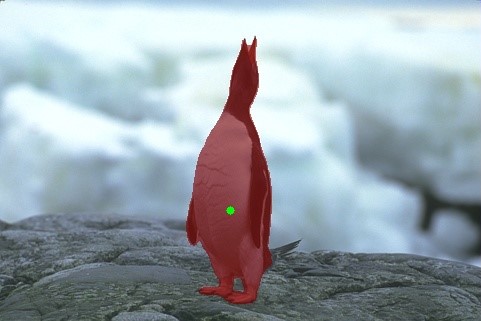}
    \subcaption*{1 click 93.2\%}
  \end{subfigure}
  \hspace{2mm}
  \begin{subfigure}[t]{0.22\linewidth}
    \includegraphics[width=1\linewidth]{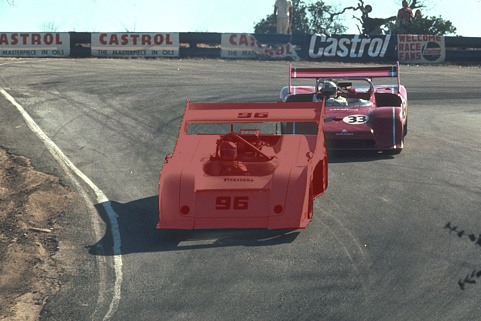}
    \subcaption*{GT}
  \end{subfigure}
  \hspace{2mm}
  \begin{subfigure}[t]{0.22\linewidth}
    \includegraphics[width=1\linewidth]{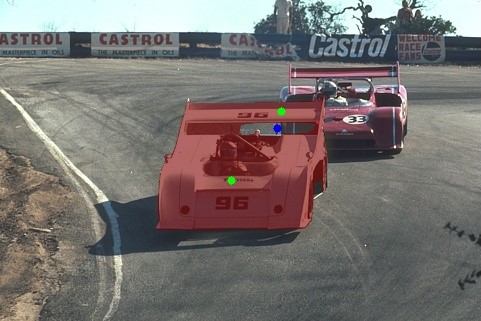}
    \subcaption*{3 clicks 92.7\%}
  \end{subfigure}
  
  \begin{subfigure}[t]{0.22\linewidth}
    \includegraphics[width=1\linewidth]{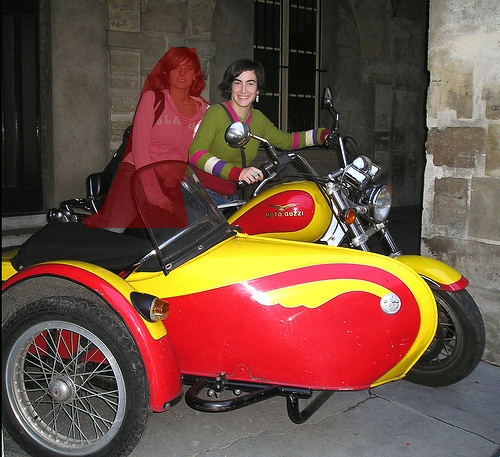}
    \subcaption*{GT}
  \end{subfigure}
  \hspace{2mm}
  \begin{subfigure}[t]{0.22\linewidth}
    \includegraphics[width=1\linewidth]{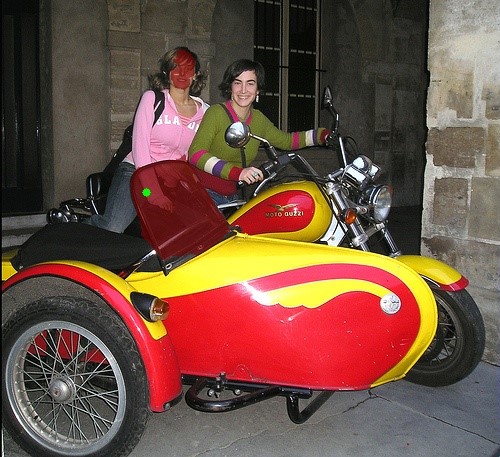}
    \subcaption*{1 click 25.6\%}
  \end{subfigure}
  \hspace{2mm}
  \begin{subfigure}[t]{0.22\linewidth}
    \includegraphics[width=1\linewidth]{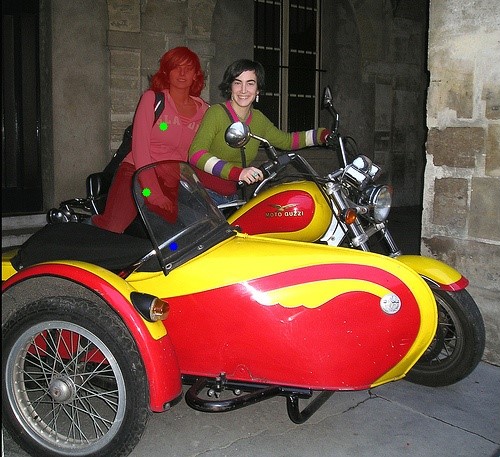}
    \subcaption*{3 clicks 80.5\%}
  \end{subfigure}
  \hspace{2mm}
  \begin{subfigure}[t]{0.22\linewidth}
    \includegraphics[width=1\linewidth]{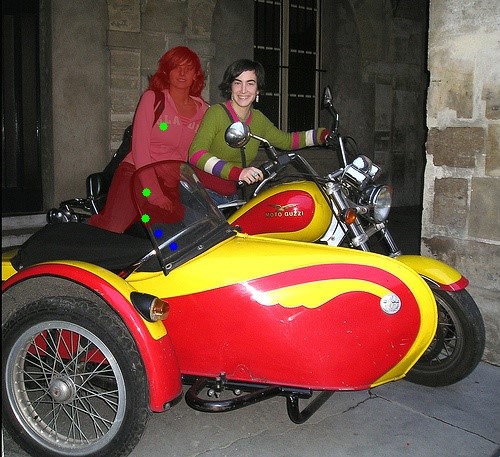}
    \subcaption*{5 clicks 90.5\%}
  \end{subfigure}

  \begin{subfigure}[t]{0.22\linewidth}
    \includegraphics[width=1\linewidth]{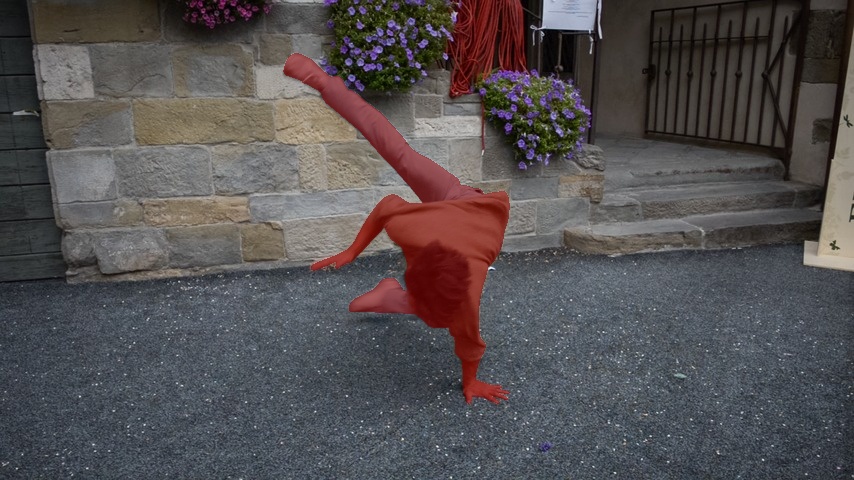}
    \subcaption*{GT}
  \end{subfigure}
  \hspace{2mm}
  \begin{subfigure}[t]{0.22\linewidth}
    \includegraphics[width=1\linewidth]{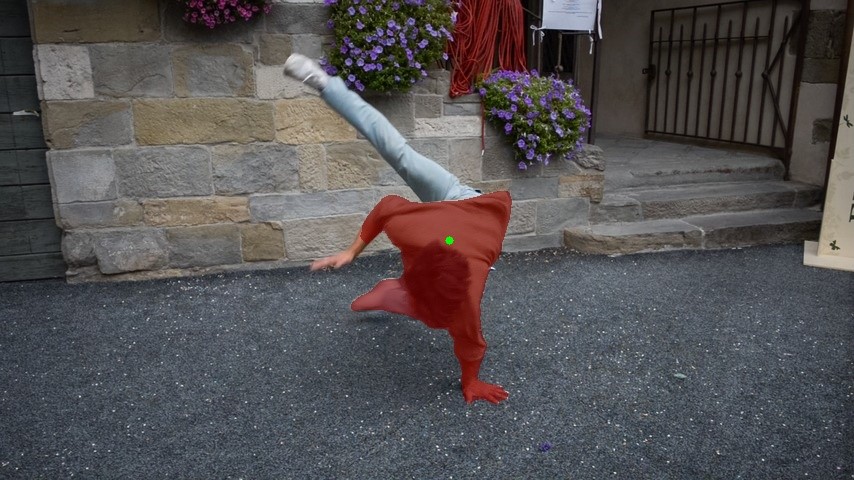}
    \subcaption*{1 click 66.3\%}
  \end{subfigure}
  \hspace{2mm}
  \begin{subfigure}[t]{0.22\linewidth}
    \includegraphics[width=1\linewidth]{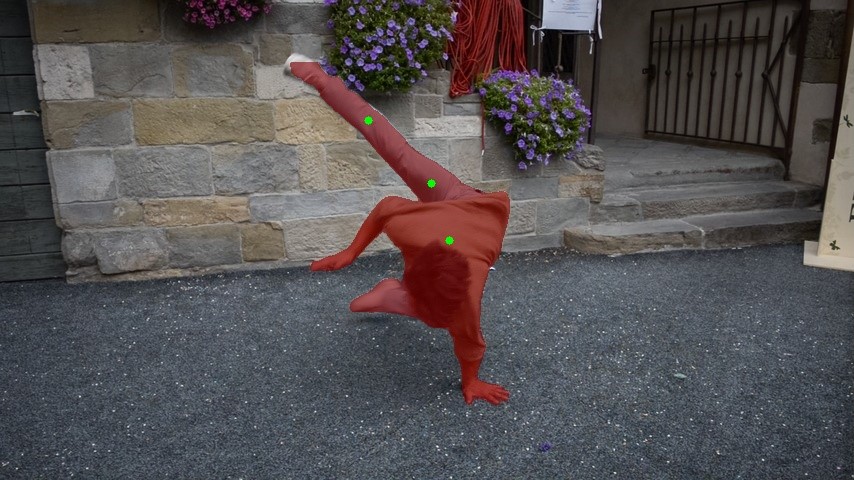}
    \subcaption*{3 clicks 93.3\%}
  \end{subfigure}
  \hspace{2mm}
  \begin{subfigure}[t]{0.22\linewidth}
    \includegraphics[width=1\linewidth]{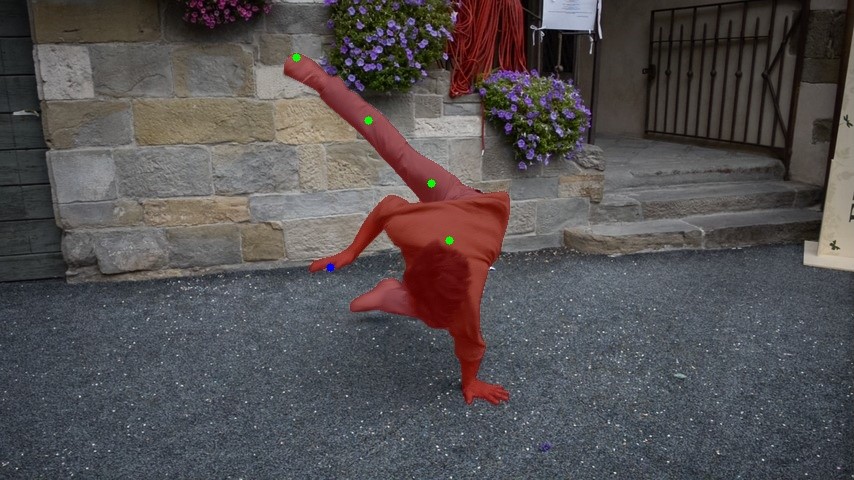}
    \subcaption*{5 clicks 95.4\%}
  \end{subfigure}

  \begin{subfigure}[t]{0.22\linewidth}
    \includegraphics[width=1\linewidth]{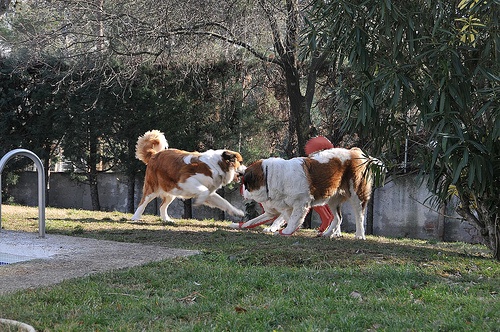}
    \subcaption*{GT}
  \end{subfigure}
  \hspace{2mm}
  \begin{subfigure}[t]{0.22\linewidth}
    \includegraphics[width=1\linewidth]{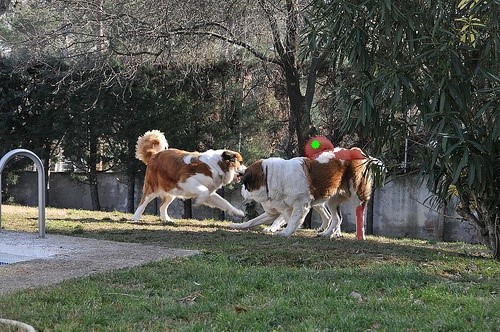}
    \subcaption*{1 click 21.4\%}
  \end{subfigure}
  \hspace{2mm}
  \begin{subfigure}[t]{0.22\linewidth}
    \includegraphics[width=1\linewidth]{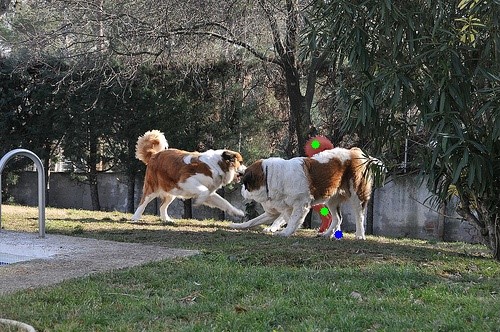}
    \subcaption*{3 clicks 59.8\%}
  \end{subfigure}
  \hspace{2mm}
  \begin{subfigure}[t]{0.22\linewidth}
    \includegraphics[width=1\linewidth]{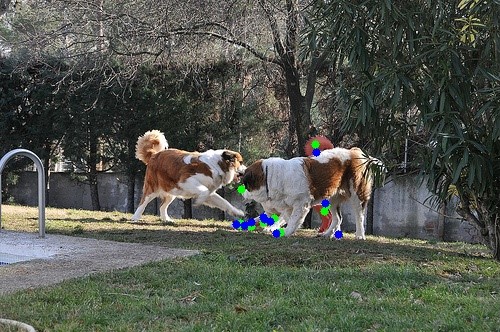}
    \subcaption*{20 clicks 68.0\%}
  \end{subfigure}
  \vspace{-3mm}
  \caption{Visualizations of our segmentation results. The segmentation results are displayed in masks, and the corresponding IoU values with different clicks. Green and blue dots denote positive and negative clicks, respectively. Row 1-3 display some successful cases from the four datasets while the last row shows a bad case from SBD \cite{hariharan2011sbd}.}
  \vspace{-5mm}
  \label{fig:segmentations}
\end{figure*}

\noindent
\textbf{Performance on Benchmarks.}
The comparison results on GrabCut \cite{rother2004grabcut}, Berkeley \cite{martin2001berkely}, SBD \cite{hariharan2011sbd}, and DAVIS \cite{perazzi2016davis} with respect to the number of clicks (NoC) are demonstrated in Tab.~\ref{tab:comparison with SOTA}. As some of the methods are trained in different datasets (early on PASCAL \cite{everingham2009pascal}, popularly on SBD, and recently on COCO \cite{lin2014coco} + LVIS \cite{gupta2019lvis}), we split the table into 3 sections. We also report the backbones of different methods to indicate the importance of feature extraction. Our proposed iCMFormer reaches the state-of-the-art on 4 datasets when trained on SBD. For instance, on DAVIS (a high-quality gold standard of ground truths), it succeeds in reducing almost one click required to reach the higher IoU threshold. Additionally, our iCMFormer achieves competitive results when trained on COCO + LVIS. It significantly improves the results on Berkeley, achieving 90\% IoU with less than 2 clicks, and sets the new state-of-the-arts on highly competitive benchmarks such as SBD and DAVIS. The results surpass previous methods and demonstrate the effectiveness of our proposed method.

To visually compare the segmentation performance with other methods, Fig.~\ref{fig:anaysis} illustrates the mean IoU curves with progressively added clicks on SBD and DAVIS datasets. Due to the limited space, the curves of the other two datasets are shown in the supplementary material. We can observe that our methods achieve better mean IoU scores with the same number of clicks, and require fewer clicks to reach the same target IoU. For instance, ours-Swin-B improves the mIoU performance to around 75\% with only one click on SBD. The figures also proves the superiority of our method to others shown in Tab.~\ref{tab:comparison with SOTA} when analysing the first 5 clicks.

As a practical annotation tool, it is extremely necessary and vital to obtain high-quality segmentation masks of targets if provided with sufficient clicks. Then we report the number of failures (NoF$\geq$20@90) for 3 datasets on Tab.~\ref{tab:comparison with SOTA for nums larger than 20} (more complex compared to GrabCut). The proposed iCMFormer improves the results on the 3 datasets compared with the others. Remarkably, it reduces the failure cases below 700 on SBD, which outperforms the previous refinement method f-BRS \cite{sofiiuk2020fbrs} by 52.7\%. Note that we only report the numbers that are provided by the original papers and their released pre-trained models. Limited by the space, we provide more details in the supplementary material.

\noindent
\textbf{Computation Analysis.}
We perform the computation analysis in terms of parameters, FLOPs, and inference speed. In Tab.~\ref{tab:computation comparison}, we report the corresponding models to represent various methods. To make a fair comparison, we set the same computing environment (NVIDIA A40 GPU and Intel Silver 4216 CPU). However, some methods process input images with different sizes (e.g., FocalClick \cite{chen2022focalclick} dealing with smaller size 256 while most methods with around 400). To address this issue, we also report the image size to complement the comparison. The numbers of parameters are collected from the original works \cite{chen2022focalclick, lin2022focuscut,sofiiuk2022ritm}. Although both proposed backbones require more parameters, their inference speeds (e.g., 78ms, 86ms) still meet the requirements for real-time interactive segmentation. We also provide the numbers for a smaller variant based on Swin-S in Tab.~\ref{tab:comparison with SOTA} and Tab.~\ref{tab:computation comparison}. Our proposed end-to-end method still beats the current SOTAs with a comparable backbone.

\begin{table}
  \caption{Ablation study for different components trained on SBD \cite{hariharan2011sbd}. NoC@90 denotes the average numbers of clicks to reach 90\% IoU. The best results are \textbf{bold}.}
  \centering
  \setlength\tabcolsep{3pt}
  \begin{tabular}{ccccc}
  \toprule
  \multirow{2}*{Cross-M} & \multirow{2}*{Hierarchy} & \multicolumn{1}{c}{Berkeley} & \multicolumn{1}{c}{SBD} & \multicolumn{1}{c}{DAVIS}\\
 & & NoC@90 & NoC@90 & NoC@90\\
  \midrule
  w/o  & w/o  & 2.55 & 6.05 & 5.76\\
  w/o  & w & 2.58 & 5.57 & 5.63 \\
  w & w/o  & 2.52 & 5.31 & 5.58\\
  w  & w  & \textbf{2.32} & \textbf{5.16} & \textbf{5.55}\\
  \bottomrule
  \end{tabular}
  \label{tab:ablation study1}
  \vspace{-3mm}
\end{table}

\begin{table}
  \caption{Ablation study for the proposed ViT-B \cite{dosovitskiy2020vit} backbone with different variants trained on SBD \cite{hariharan2011sbd}. $X$ and $Y$ denote image and click features, respectively. $\overline{X}$ and $\overline{Y}$ represent the first group of self-attentions. $\overrightarrow{YX}$ means the guidance from $Y$ to $X$, vice versa. The second group of transformers ($\overline{X \oplus Y}$) are not shown here for brevity.}
  \centering
  \setlength\tabcolsep{2pt}
  \begin{tabular}{p{1.5cm}cccc}
  \toprule
  \multirow{2}*{Variants} & \multicolumn{1}{c}{GrabCut} & \multicolumn{1}{c}{Berkeley} & \multicolumn{1}{c}{SBD} & \multicolumn{1}{c}{DAVIS}\\
 & NoC@90 & NoC@90 & NoC@90 & NoC@90\\
  \midrule
  $\overline{X}, \overrightarrow{YX}$  & 2.76 & 4.82 & 8.11 & 8.40\\
  $\overline{X}, \overrightarrow{XY}$  & 1.74 & \textbf{2.47} & 5.81 & 5.60\\
  $\overline{X}, \overline{Y}, \overrightarrow{YX}$ & 1.72  & 2.60 & 5.53 & 5.80\\
  $\overline{X}, \overline{Y}, \overrightarrow{XY}$ & \textbf{1.42}  & 2.52 & \textbf{5.31} & \textbf{5.58}\\
  \bottomrule
  \end{tabular}
  \label{tab:ablation study2}
  \vspace{-3mm}
\end{table}
\subsection{Ablation Studies}
\label{ablation}
To verify the effectiveness of the proposed method, we ablate the different components and the variants of the backbones for interactive image segmentation (number of cross-modality blocks is reported in the supplementary material). Simply, we train the models on SBD \cite{hariharan2011sbd} and automatically evaluate the NoC@90 metric on the 4 datasets.


\vspace{1mm}
\noindent
\textbf{Effectiveness of Components.} We set the original plain vision transformers \cite{dosovitskiy2020vit} with two shared branches for the first group of self-attention blocks (see in Sec.~\ref{backbone}) as the base model. The proposed cross-modality transformers aim for learning the guidance signal between two branches while the hierarchical architecture addresses the multi-scale problem in the dense prediction. We then evaluate the impact of these two components individually through the ablation study, and show the results in Tab.~\ref{tab:ablation study1}. The third row highlights the efficacy of cross-modality transformers. With hierarchy, the combined version (last row) further reduces the number of clicks, especially almost one click drop compared with base model for various instances from SBD.

\vspace{2mm}
\noindent
\textbf{Holistic Analysis.} To investigate the optimal usage of the proposed cross-modality transformers, we run the holistic analysis on the backbone variants. We keep the second group of transformer the same fed by the element-wise addition input, and focus solely on the first group and the way of guidance. The results on 4 datasets are shown in Tab.~\ref{tab:ablation study2}. The first row shows that directly guiding the image feature learning with original clicks hugely hurts the performance because of the mismatched value ranges, and the third row verifies the significance of self-attentions on the click branch. Moreover, we see that $\overline{X}, \overline{Y}, \overrightarrow{XY}$ outperforms $\overline{X}, \overline{Y}, \overrightarrow{YX}$, which reveals the key role of image features for segmentation. Due to the similar group allocation, we adopt $\overline{X}, \overline{Y}, \overrightarrow{XY}$ as our default backbone architecture for both plain and hierarchical vision transformers.
\subsection{Qualitative Results}
Visualisations of the manual evaluation process with the proposed method are shown in Fig.~\ref{fig:segmentations}. The first three rows display the examples from GrabCut \cite{rother2004grabcut}, Berkeley \cite{martin2001berkely}, SBD \cite{hariharan2011sbd} and DAVIS \cite{perazzi2016davis}, respectively. These examples show that the segmentation results get better with progressive interactions on the incorrect prediction regions. The last row gives a failure case from SBD, indicating that our method cannot address the occlusion problem when the target is only partly visible. We provide more segmentation results in the supplementary material.

\begin{figure}[t!]
  \centering
  \includegraphics[height=0.4\linewidth]{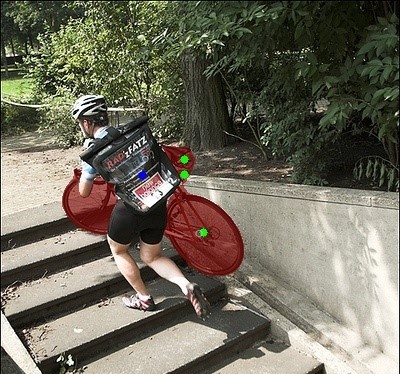}
  \includegraphics[height=0.4\linewidth]{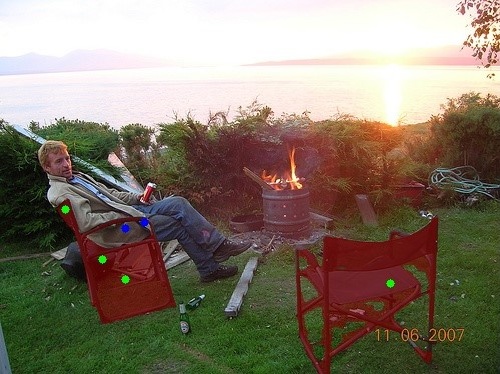}
  \caption{Examples of some disconnected region predictions from SBD \cite{hariharan2011sbd}. The left figure shows one instance with several parts, while the right illustrates multiple instances of the same category.}
  \vspace{-5mm}
  \label{fig:findings}
\end{figure}

\subsection{Discussion}
\label{discussion}
In this section, we discuss the limitations of our method and an interesting finding that emerged during the evaluation stage. As shown in the last row of Fig.~\ref{fig:segmentations}, the segmentation result is not sufficient when the target is cluttered. Fortunately, local refinements \cite{lin2022focuscut,xu2016dios} coupled with post-processing optimizations \cite{xu2017deepgrabcut} would enhance the accuracy. Given that SBD \cite{hariharan2011sbd} contains some training samples with disconnected regions, we discover that the proposed iCMFormer even learns to adapt to the interactions for different instances of the same category (in Fig.~\ref{fig:findings}). This finding can be further explored for more efficient interactive annotations in certain cases involving multiple instances.

\section{Conclusion}
In this paper, we propose a simple yet effective interactive segmentation method that leverages vision transformers. To explore the modality guidance between images and clicks for improving the accuracy of dense predictions, we raise cross-modality attentions by embedding them into both plain and hierarchical vision transformers, yielding high-quality and robust masks. The experiments demonstrate that our method achieves the best performances over four mainstream interactive segmentation datasets.
\newpage

\appendix
\section*{Appendix}
In this supplementary document, we provide detailed explanation on the architecture of the proposed iCMFormer in Sec.~\ref{sec:implementation}. 
Additional quantitative results in terms of the mIoU curves and number of failures are provided in Sec.~\ref{nof}, together with an ablation study on the number of cross-modality blocks in Sec~\ref{sec. ablation study on number of layer}. 
Moreover, we also provide more qualitative results evaluated on the 4 datasets in Sec.~\ref{sec: more vis}.

\section{Implementation Details}
\label{sec:implementation}
In the main paper, we explain the overall pipeline of the proposed iCMFormer for 2 different backbones. For better readability and reproducibility, we present the architecture in detail. As the transformer technique is quite popular, we do not expand the multi-head attentions for each block, and only report the dimension as well as the number of corresponding heads. Our iCMFormer for ViT-B and Swin-B backbones are showsn in Tab.~\ref{tab:architecture}.

\begin{figure}[ht]
  \centering
  \begin{subfigure}[t]{0.9\linewidth}
    \includegraphics[trim= 14pt 0pt 14pt 14pt, clip=True, width=1\linewidth]{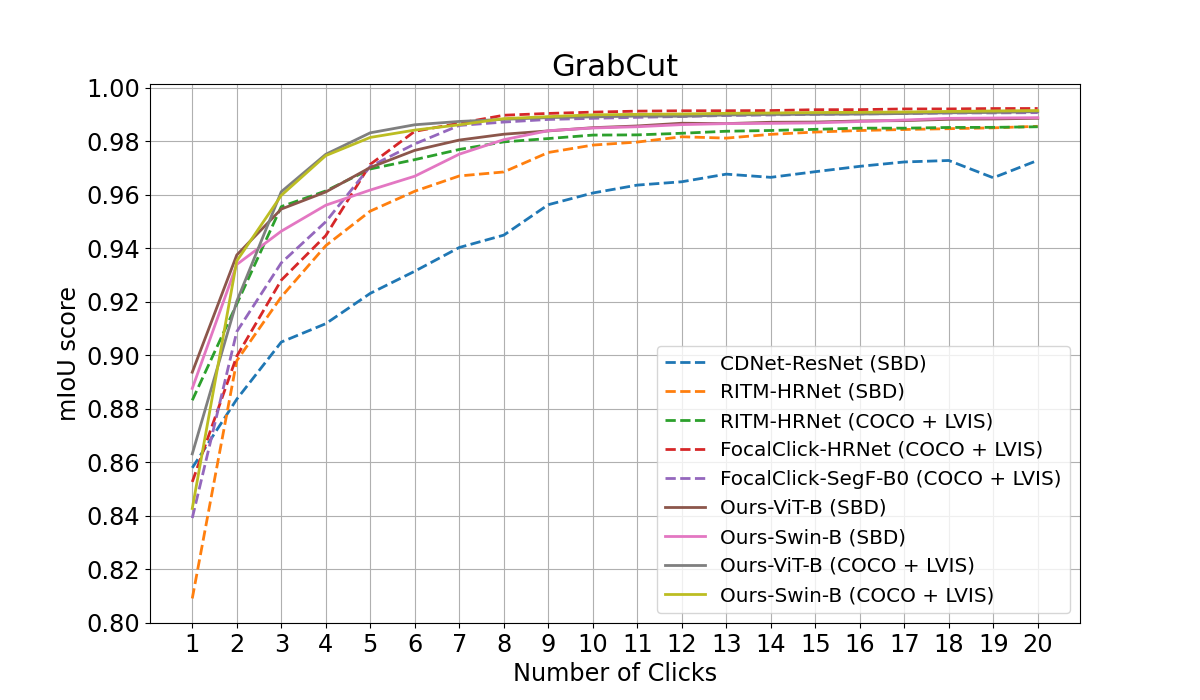}
  \end{subfigure}
  \begin{subfigure}[t]{0.9\linewidth}
    \includegraphics[trim= 14pt 0pt 14pt 14pt, clip=True, width=1\linewidth]{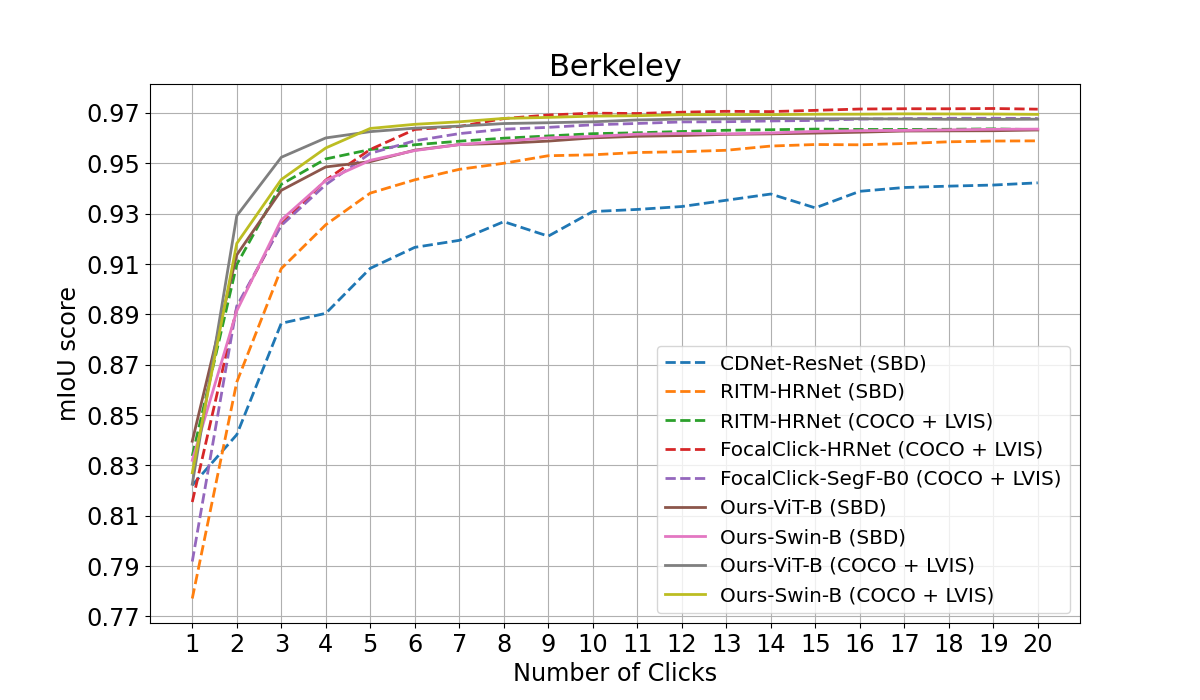}
  \end{subfigure}
  \caption{Convergence analysis of mean IoU curves for varying number of clicks compared with other methods on GrabCut \cite{rother2004grabcut} and Berkeley \cite{martin2001berkely}.}
  \label{fig:anaysis easy 2 datasets}
\end{figure}

\begin{table*}
  \caption{The detailed architecture of the iCMFormer with ViT-B and Swin-B backbones. Numbers in square brackets [] mean the input and hidden dimensions, respetively, while the numbers in parentheses () denote the dimension changes in the Conv2d or ConvTranspose2d (only utilized in the Neck) or Linear operations. We set 8 as the numbers of heads for all blocks in ViT-B, and 4, 8, 16, 32 for 4 original stages in Swin-B. The number is set 8 for all cross-attentions for both backbones.}
  \centering
  \setlength\tabcolsep{4pt}
  \begin{tabular}{p{2.3cm}|c|c|c|c}
  \hline
  Layer Name & \multicolumn{2}{c|}{Ours-ViT-B} & \multicolumn{2}{c}{Ours-Swin-B}\\
  \hline
  Patch-Embed & \multicolumn{2}{c|}{(3, 768)} & \multicolumn{2}{c}{(3, 128)}\\
  \hline
  \multirow{2}*{Shared Group} &
  \multirow{2}*{[768, 2304] + (768, 3072, 768)} & \multirow{2}*{x6} & \multicolumn{1}{|c}{[128, 384] + (128, 512, 128)} & \multicolumn{1}{|c}{x2}\\
  \cline{4-5}
 & & & \multicolumn{1}{|c}{[256, 768] + (256, 1024, 256)} & \multicolumn{1}{|c}{x2} \\
 \hline
 Cross-Attention &
  [768, 2304] + [768, 2304] + (768, 3072, 768) & x3 & [512, 1536] + [512, 1536] + (512, 2048, 512) & x4\\
 \hline
 \multirow{2}*{Self-Attention} &
  \multirow{2}*{[768, 2304] + (768, 3072, 768)} & \multirow{2}*{x6} & \multicolumn{1}{|c}{[512, 1536] + (512, 2048, 512)} & \multicolumn{1}{|c}{x18}\\
  \cline{4-5}
 & & & \multicolumn{1}{|c}{[1024, 3072] + (1024, 4096, 1024)} & \multicolumn{1}{|c}{x2}\\
 \hline
 \multirow{4}*{Neck} & \multicolumn{2}{c|}{(768, 384, 192, 128)} & \multicolumn{2}{c}{\multirow{4}*{-}}\\
 & \multicolumn{2}{c|}{(768, 384, 256)}\\
 & \multicolumn{2}{c|}{(768, 512)}\\
 & \multicolumn{2}{c|}{(768, 1536, 1024)}\\
 \hline
 \multirow{5}*{Head} & \multicolumn{2}{c|}{(128, 256)} & \multicolumn{2}{c}{(128, 256)}\\
 & \multicolumn{2}{c|}{(256, 256)} & \multicolumn{2}{c}{(256, 256)}\\
 & \multicolumn{2}{c|}{(512, 256)} & \multicolumn{2}{c}{(512, 256)}\\
 & \multicolumn{2}{c|}{(1024, 256)} & \multicolumn{2}{c}{(1024, 256)}\\
 & \multicolumn{2}{c|}{(256$\times$4, 256, 1)} & \multicolumn{2}{c}{(256$\times$4, 256, 1)}\\
  \hline
  \end{tabular}
  \label{tab:architecture}
\end{table*}


\section{Additional Quantitative Results}
\label{nof}
In the main paper, we report the complete comparison results with respect to the Number of Clicks (NoC). Due to the limited space, here we further provide the evaluation results in terms of mean IoU curves and Number of Failures (NoF) to make the comparison consistent with the employed evaluation protocol.

We report the automatically evaluation results on GrabCut \cite{rother2004grabcut} and Berkeley \cite{martin2001berkely} in Fig.~\ref{fig:anaysis easy 2 datasets} for demonstrating the segmentation performance with progressively added clicks. We can see that the proposed methods achieve higher mIoU values within the same number of clicks compared with other models. However, restricted in the sizes of evaluation samples in GrabCut (50) and Berkeley (100), different variants of our methods do not make a huge difference especially when only providing 2 clicks (already above 90\% mIoU).

In addition, we compared the stability of our method with that of others in Tab.~\ref{tab:failure 2 datasets} using 20 clicks for 2 thresholds: 85\% and 90\%. As the previous methods did not report the numbers for GrabCut and Berkeley, we do not add the values in the table (Ours-Swin-B only gets both 0 failure on GrabCut and 0, 1 failure on Berkeley for 85\% and 90\% IoU, respectively). The models trained in SBD \cite{hariharan2011sbd} and COCO \cite{lin2014coco} + LVIS \cite{gupta2019lvis} are divided into 2 parts for fair comparison. As shown in the table, our method reduces the numbers of failure cases for both thresholds, which show the potential to be a practical annotation tool with robust predictions.

\section{Number of Cross-Modality Blocks}
\label{sec. ablation study on number of layer}
We further evaluate the impact of different number of the proposed cross-modality blocks on the performance of our backbones. Simply, we train all the models on SBD \cite{hariharan2011sbd} and evaluate the results on 4 datasets with the NoC metric. Tab.~\ref{tab:ablation study3} shows the corresponding results. As the number of layers increases, the trend of the number of clicks (NoC) shows an initial rise followed by a subsequent decline. Due to the better overall performance, we set 3 and 4 as the default numbers for ViT-B and Swin-B backbones, respectively.

\section{More Qualitative Results}
\label{sec: more vis}
\begin{figure*}[ht]
  \centering
  \begin{subfigure}[t]{0.22\linewidth}
    \includegraphics[width=1\linewidth]{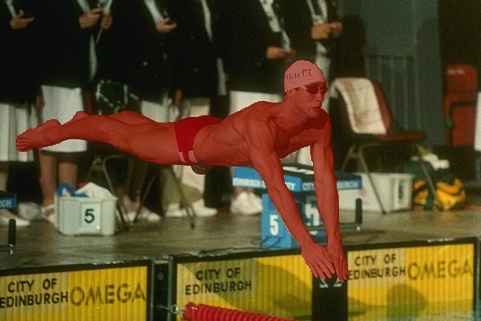}
    \subcaption*{GT}
  \end{subfigure}
  \hspace{2mm}
  \begin{subfigure}[t]{0.22\linewidth}
    \includegraphics[width=1\linewidth]{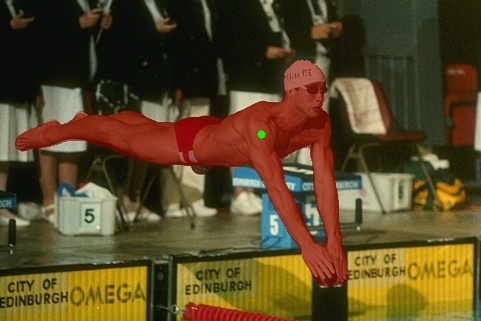}
    \subcaption*{1 click 93.4\%}
  \end{subfigure}
  \hspace{2mm}
  \begin{subfigure}[t]{0.22\linewidth}
    \includegraphics[width=1\linewidth]{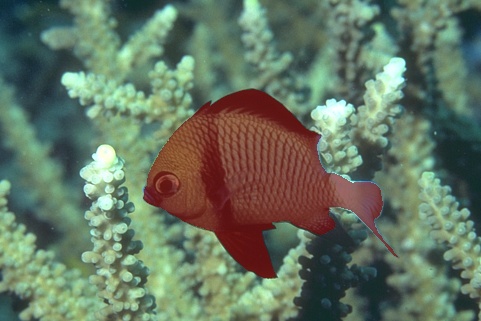}
    \subcaption*{GT}
  \end{subfigure}
  \hspace{2mm}
  \begin{subfigure}[t]{0.22\linewidth}
    \includegraphics[width=1\linewidth]{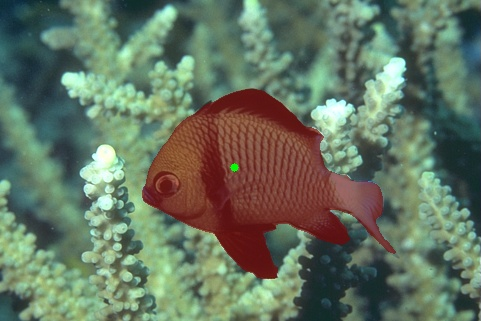}
    \subcaption*{1 click 92.4\%}
  \end{subfigure}

  \begin{subfigure}[t]{0.22\linewidth}
    \includegraphics[width=1\linewidth]{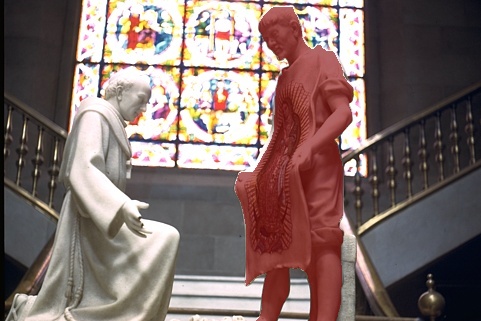}
    \subcaption*{GT}
  \end{subfigure}
  \hspace{2mm}
  \begin{subfigure}[t]{0.22\linewidth}
    \includegraphics[width=1\linewidth]{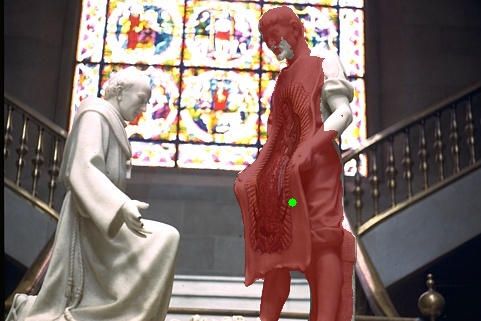}
    \subcaption*{1 click 81.3\%}
  \end{subfigure}
  \hspace{2mm}
  \begin{subfigure}[t]{0.22\linewidth}
    \includegraphics[width=1\linewidth]{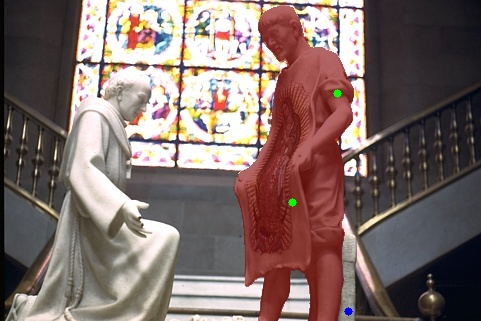}
    \subcaption*{3 clicks 94.1\%}
  \end{subfigure}
  \hspace{2mm}
  \begin{subfigure}[t]{0.22\linewidth}
    \includegraphics[width=1\linewidth]{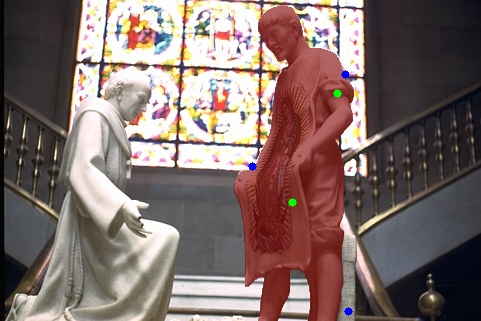}
    \subcaption*{5 clicks 95.1\%}
  \end{subfigure}

  \begin{subfigure}[t]{0.22\linewidth}
    \includegraphics[width=1\linewidth]{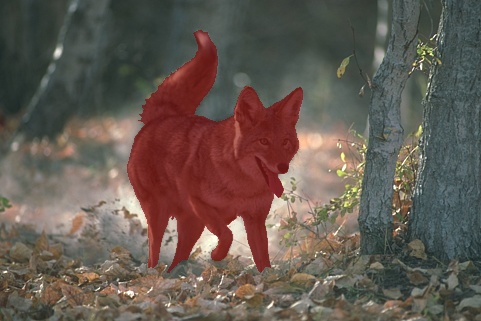}
    \subcaption*{GT}
  \end{subfigure}
  \hspace{2mm}
  \begin{subfigure}[t]{0.22\linewidth}
    \includegraphics[width=1\linewidth]{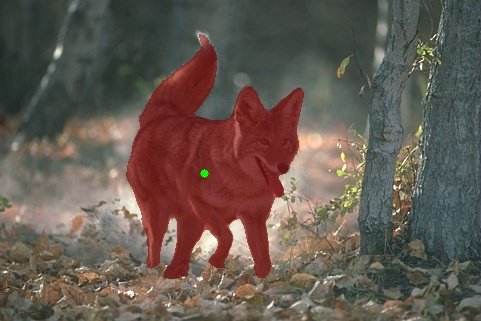}
    \subcaption*{1 click 93.2\%}
  \end{subfigure}
  \hspace{2mm}
  \begin{subfigure}[t]{0.22\linewidth}
    \includegraphics[width=1\linewidth]{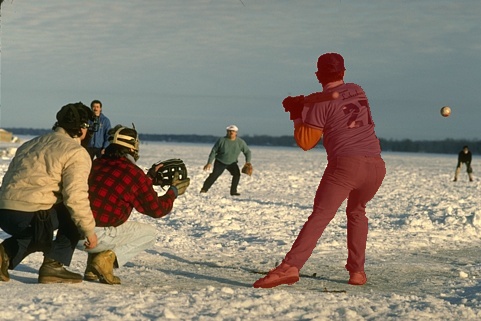}
    \subcaption*{GT}
  \end{subfigure}
  \hspace{2mm}
  \begin{subfigure}[t]{0.22\linewidth}
    \includegraphics[width=1\linewidth]{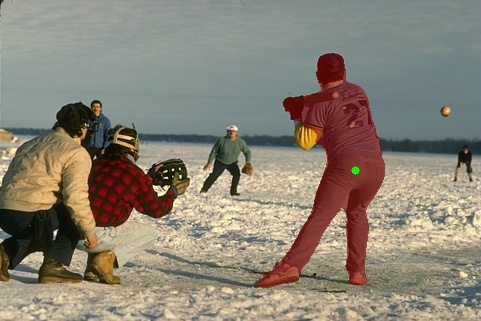}
    \subcaption*{1 click 90.6\%}
  \end{subfigure}

  \begin{subfigure}[t]{0.22\linewidth}
    \includegraphics[width=1\linewidth]{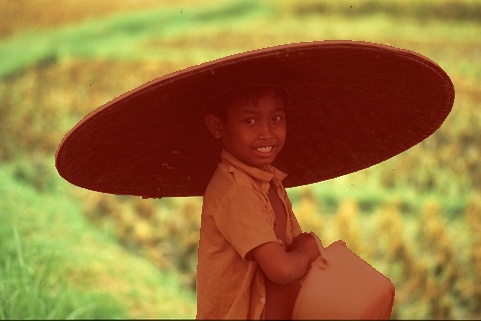}
    \subcaption*{GT}
  \end{subfigure}
  \hspace{2mm}
  \begin{subfigure}[t]{0.22\linewidth}
    \includegraphics[width=1\linewidth]{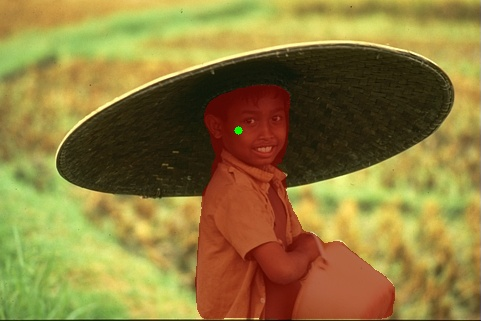}
    \subcaption*{1 click 41.4\%}
  \end{subfigure}
  \hspace{2mm}
  \begin{subfigure}[t]{0.22\linewidth}
    \includegraphics[width=1\linewidth]{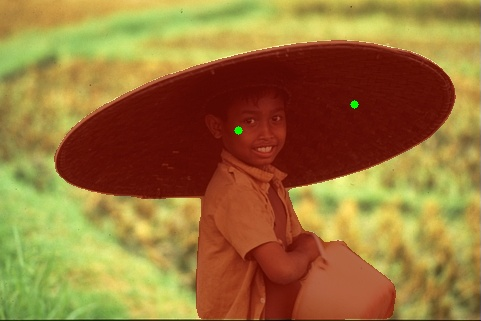}
    \subcaption*{2 clicks 98.2\%}
  \end{subfigure}
  \hspace{2mm}
  \begin{subfigure}[t]{0.22\linewidth}
    \includegraphics[width=1\linewidth]{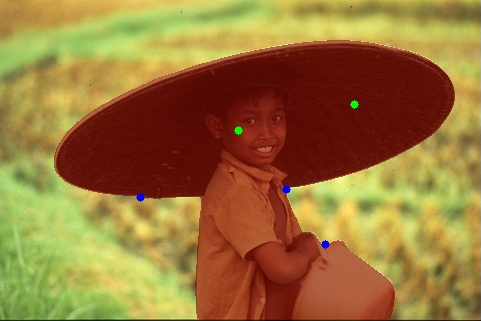}
    \subcaption*{5 clicks 98.8\%}
  \end{subfigure}

  \begin{subfigure}[t]{0.22\linewidth}
    \includegraphics[width=1\linewidth]{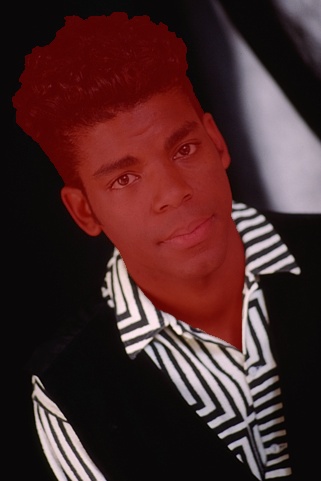}
    \subcaption*{GT}
  \end{subfigure}
  \hspace{2mm}
  \begin{subfigure}[t]{0.22\linewidth}
    \includegraphics[width=1\linewidth]{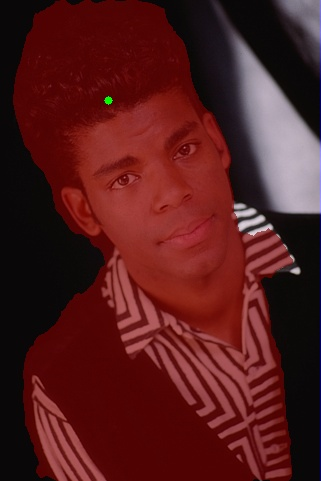}
    \subcaption*{1 click 47.2\%}
  \end{subfigure}
  \hspace{2mm}
  \begin{subfigure}[t]{0.22\linewidth}
    \includegraphics[width=1\linewidth]{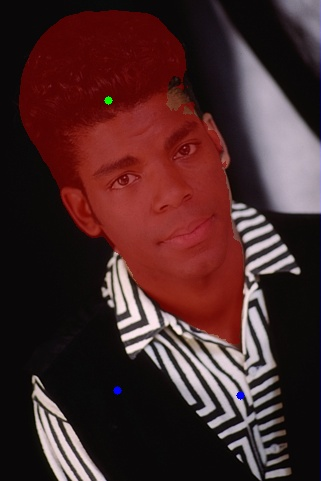}
    \subcaption*{3 clicks 92.6\%}
  \end{subfigure}
  \hspace{2mm}
  \begin{subfigure}[t]{0.22\linewidth}
    \includegraphics[width=1\linewidth]{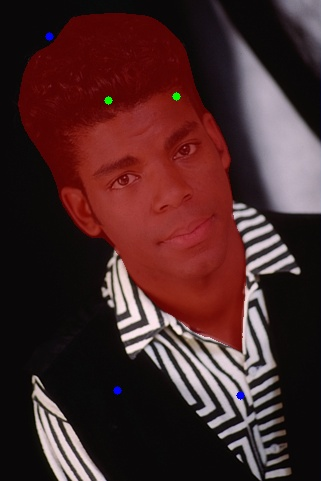}
    \subcaption*{5 clicks 95.7\%}
  \end{subfigure}
  \vspace{-2mm}
  \caption{More visualizations of the segmentation results from GrabCut \cite{rother2004grabcut} (Row 1-2) and Berkeley \cite{martin2001berkely} (Row 3-5). Green and blue dots denote positive and negative clicks, respectively.}
  \vspace{-3mm}
  \label{fig:segmentations more-grabcut-berkeley}
\end{figure*}

\begin{figure*}[ht]
  \centering
  \begin{subfigure}[t]{0.22\linewidth}
    \includegraphics[width=1\linewidth]{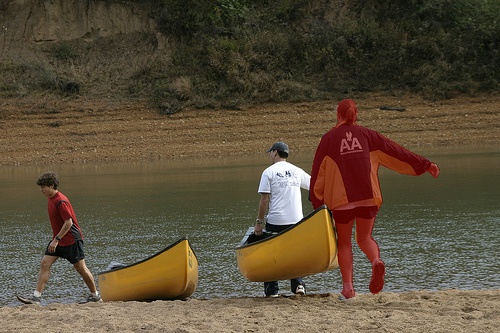}
    \subcaption*{GT}
  \end{subfigure}
  \hspace{2mm}
  \begin{subfigure}[t]{0.22\linewidth}
    \includegraphics[width=1\linewidth]{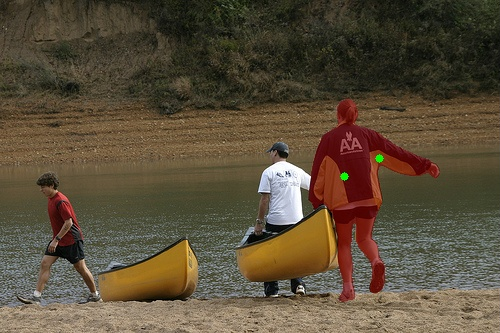}
    \subcaption*{2 clicks 95.9\%}
  \end{subfigure}
  \hspace{2mm}
  \begin{subfigure}[t]{0.22\linewidth}
    \includegraphics[width=1\linewidth]{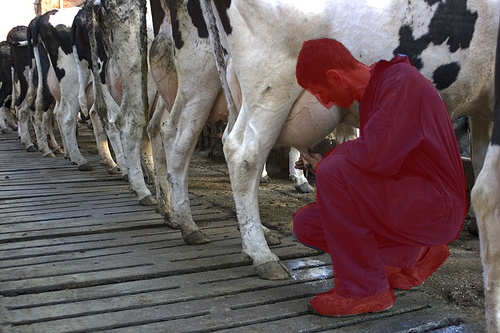}
    \subcaption*{GT}
  \end{subfigure}
  \hspace{2mm}
  \begin{subfigure}[t]{0.22\linewidth}
    \includegraphics[width=1\linewidth]{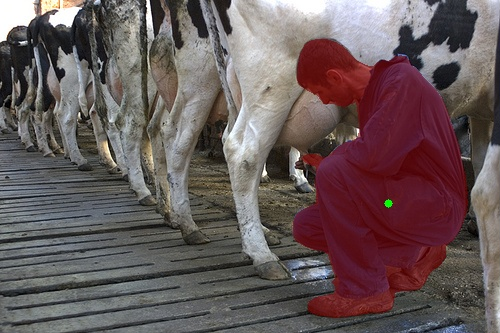}
    \subcaption*{1 click 96.0\%}
  \end{subfigure}

  \begin{subfigure}[t]{0.22\linewidth}
    \includegraphics[width=1\linewidth]{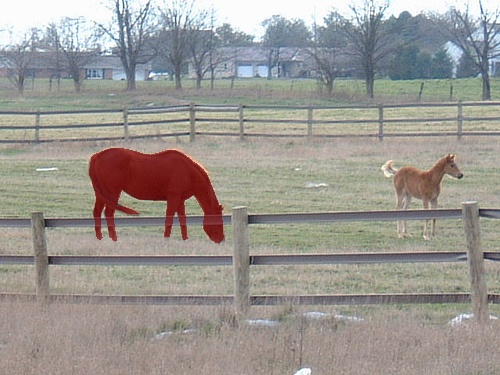}
    \subcaption*{GT}
  \end{subfigure}
  \hspace{2mm}
  \begin{subfigure}[t]{0.22\linewidth}
    \includegraphics[width=1\linewidth]{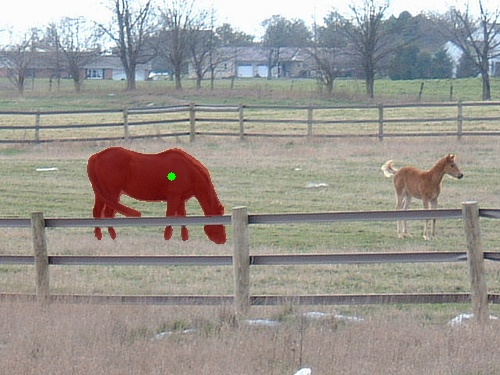}
    \subcaption*{1 click 85.4\%}
  \end{subfigure}
  \hspace{2mm}
  \begin{subfigure}[t]{0.22\linewidth}
    \includegraphics[width=1\linewidth]{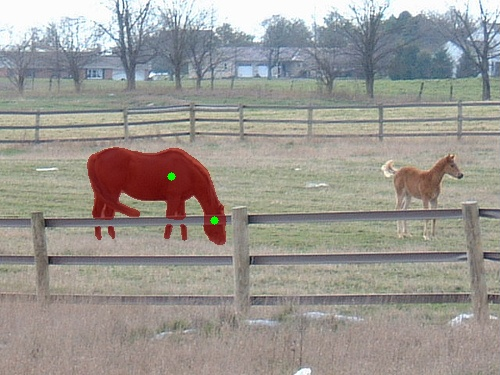}
    \subcaption*{3 clicks 87.9\%}
  \end{subfigure}
  \hspace{2mm}
  \begin{subfigure}[t]{0.22\linewidth}
    \includegraphics[width=1\linewidth]{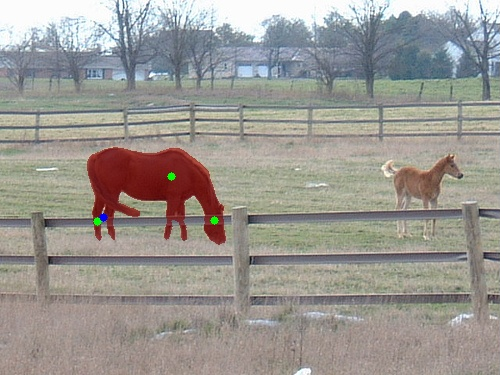}
    \subcaption*{5 clicks 90.6\%}
  \end{subfigure}
  
  \begin{subfigure}[t]{0.22\linewidth}
    \includegraphics[width=1\linewidth]{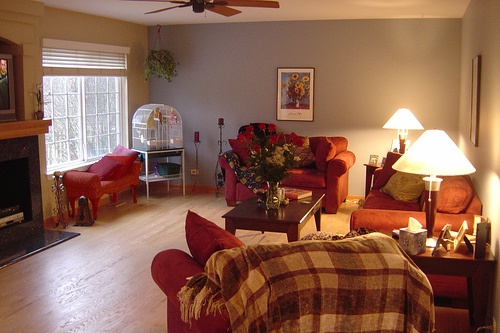}
    \subcaption*{GT}
  \end{subfigure}
  \hspace{2mm}
  \begin{subfigure}[t]{0.22\linewidth}
    \includegraphics[width=1\linewidth]{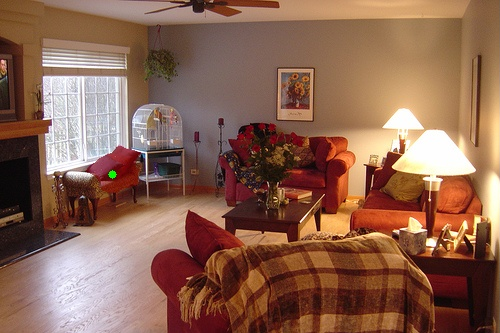}
    \subcaption*{1 click 48.3\%}
  \end{subfigure}
  \hspace{2mm}
  \begin{subfigure}[t]{0.22\linewidth}
    \includegraphics[width=1\linewidth]{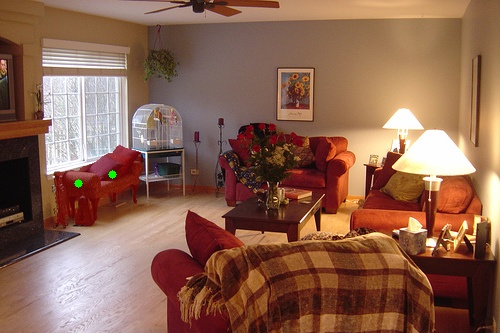}
    \subcaption*{3 clicks 63.9\%}
  \end{subfigure}
  \hspace{2mm}
  \begin{subfigure}[t]{0.22\linewidth}
    \includegraphics[width=1\linewidth]{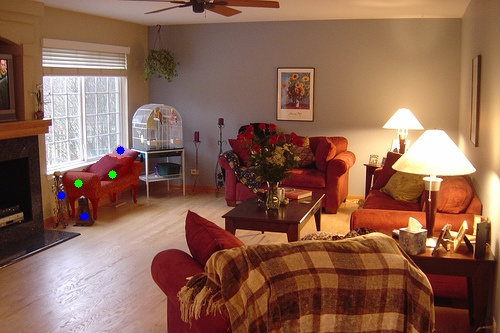}
    \subcaption*{5 clicks 91.5\%}
  \end{subfigure}

  \begin{subfigure}[t]{0.22\linewidth}
    \includegraphics[width=1\linewidth]{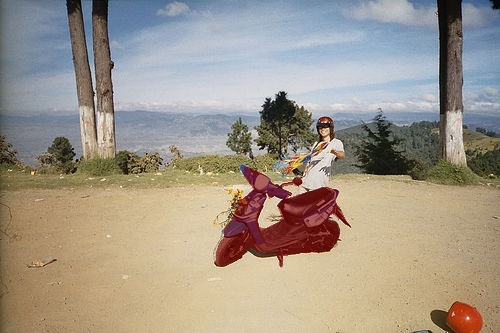}
    \subcaption*{GT}
  \end{subfigure}
  \hspace{2mm}
  \begin{subfigure}[t]{0.22\linewidth}
    \includegraphics[width=1\linewidth]{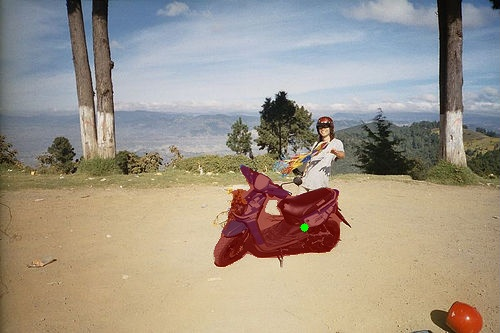}
    \subcaption*{1 click 82.3\%}
  \end{subfigure}
  \hspace{2mm}
  \begin{subfigure}[t]{0.22\linewidth}
    \includegraphics[width=1\linewidth]{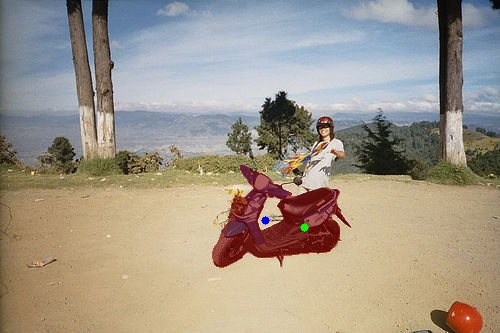}
    \subcaption*{3 clicks 87.1\%}
  \end{subfigure}
  \hspace{2mm}
  \begin{subfigure}[t]{0.22\linewidth}
    \includegraphics[width=1\linewidth]{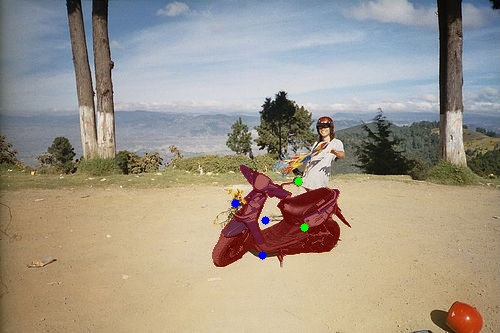}
    \subcaption*{5 clicks 90.9\%}
  \end{subfigure}

  \begin{subfigure}[t]{0.22\linewidth}
    \includegraphics[width=1\linewidth]{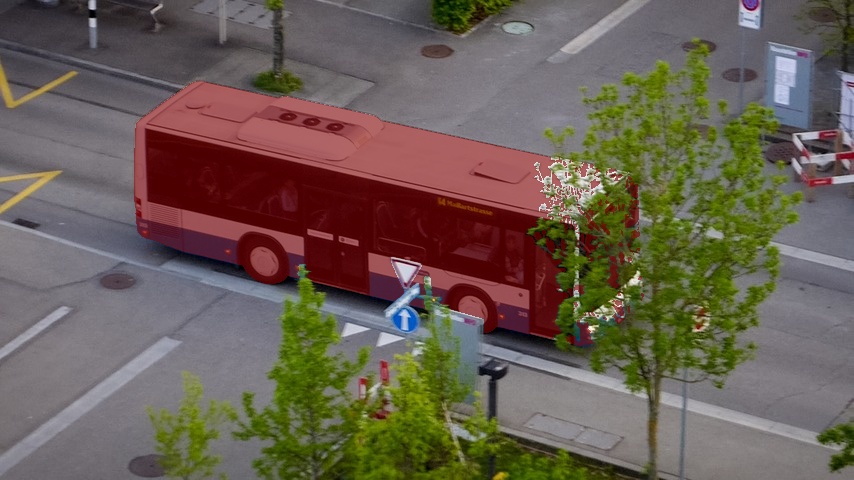}
    \subcaption*{GT}
  \end{subfigure}
  \hspace{2mm}
  \begin{subfigure}[t]{0.22\linewidth}
    \includegraphics[width=1\linewidth]{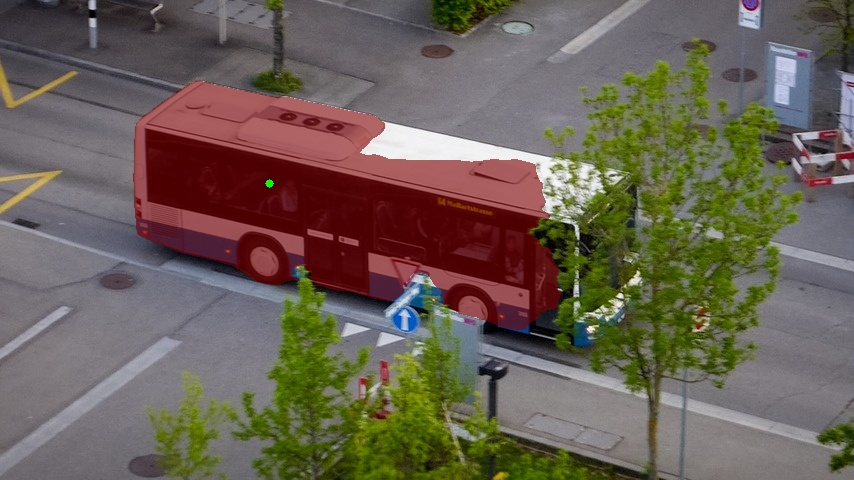}
    \subcaption*{1 click 86.4\%}
  \end{subfigure}
  \hspace{2mm}
  \begin{subfigure}[t]{0.22\linewidth}
    \includegraphics[width=1\linewidth]{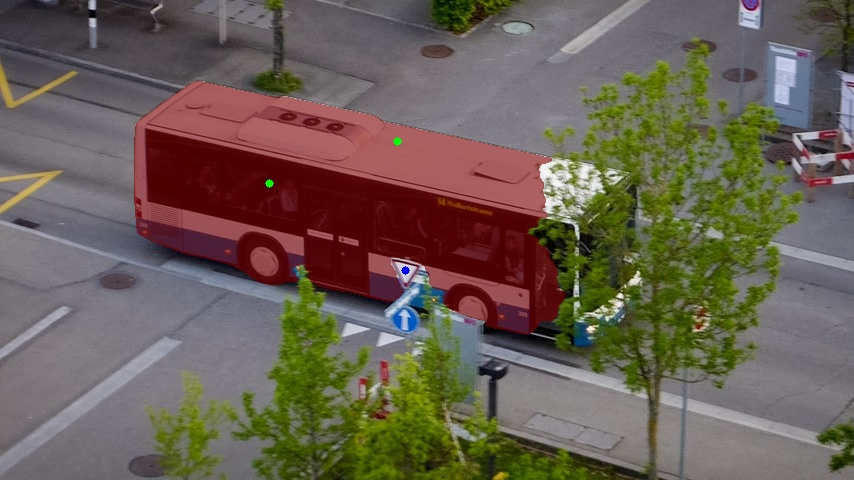}
    \subcaption*{3 clicks 92.4\%}
  \end{subfigure}
  \hspace{2mm}
  \begin{subfigure}[t]{0.22\linewidth}
    \includegraphics[width=1\linewidth]{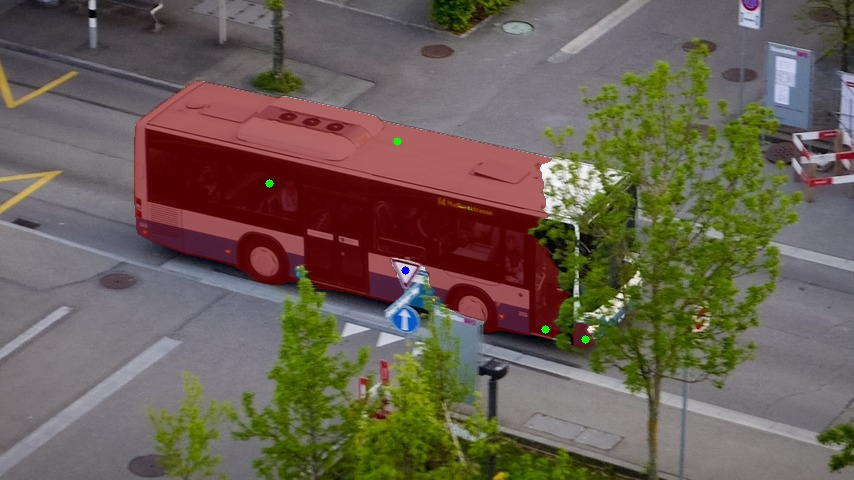}
    \subcaption*{5 clicks 93.0\%}
  \end{subfigure}

  \begin{subfigure}[t]{0.22\linewidth}
    \includegraphics[width=1\linewidth]{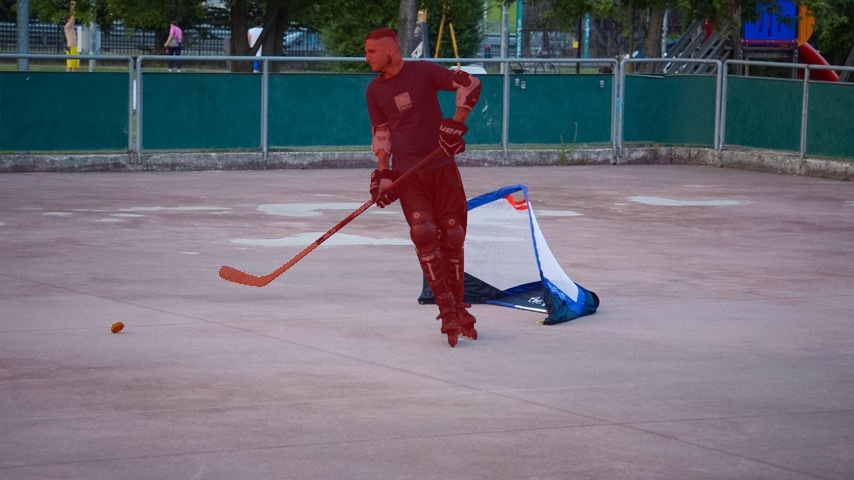}
    \subcaption*{GT}
  \end{subfigure}
  \hspace{2mm}
  \begin{subfigure}[t]{0.22\linewidth}
    \includegraphics[width=1\linewidth]{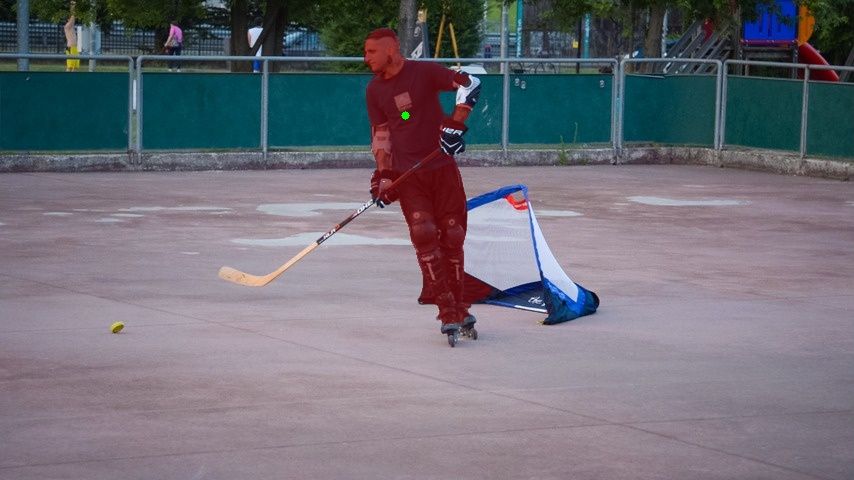}
    \subcaption*{1 click 78.8\%}
  \end{subfigure}
  \hspace{2mm}
  \begin{subfigure}[t]{0.22\linewidth}
    \includegraphics[width=1\linewidth]{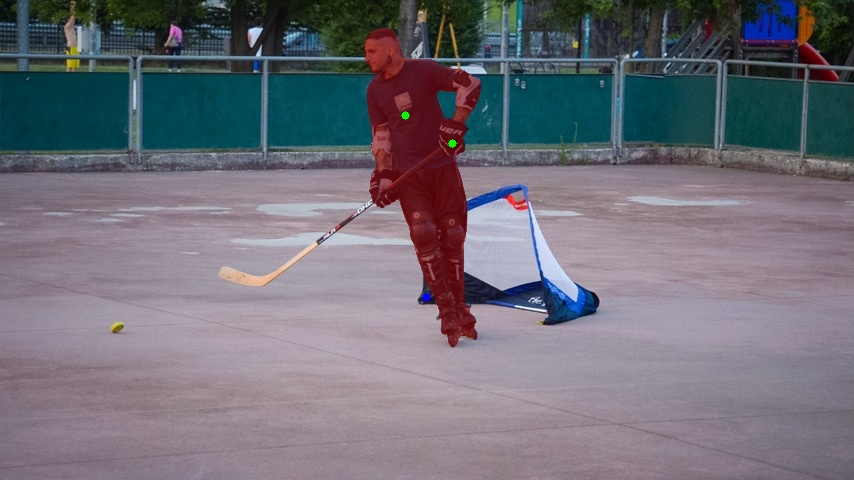}
    \subcaption*{3 clicks 90.2\%}
  \end{subfigure}
  \hspace{2mm}
  \begin{subfigure}[t]{0.22\linewidth}
    \includegraphics[width=1\linewidth]{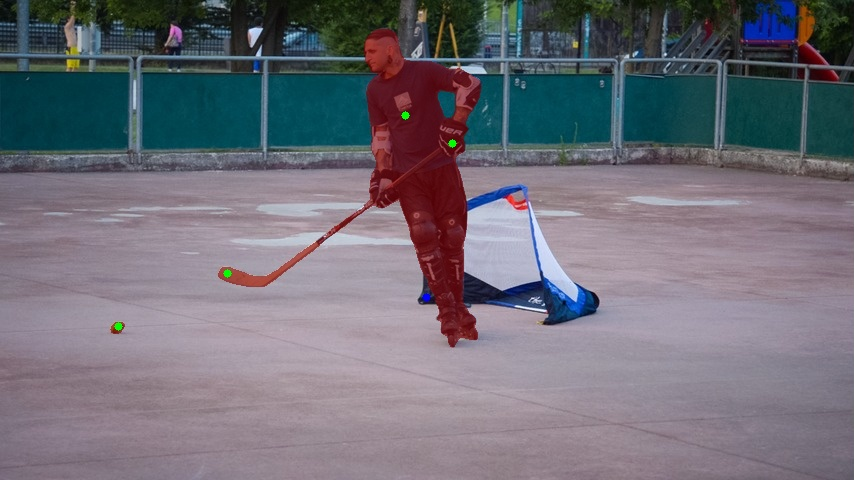}
    \subcaption*{5 clicks 94.9\%}
  \end{subfigure}
  
  \begin{subfigure}[t]{0.22\linewidth}
    \includegraphics[width=1\linewidth]{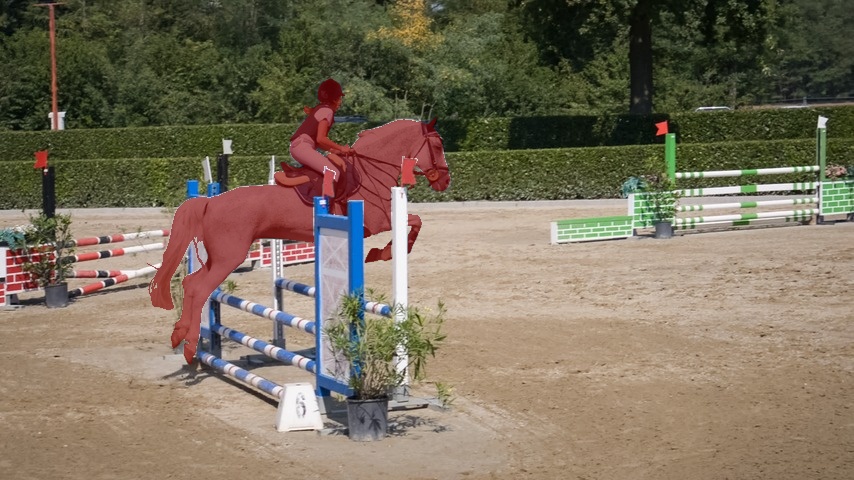}
    \subcaption*{GT}
  \end{subfigure}
  \hspace{2mm}
  \begin{subfigure}[t]{0.22\linewidth}
    \includegraphics[width=1\linewidth]{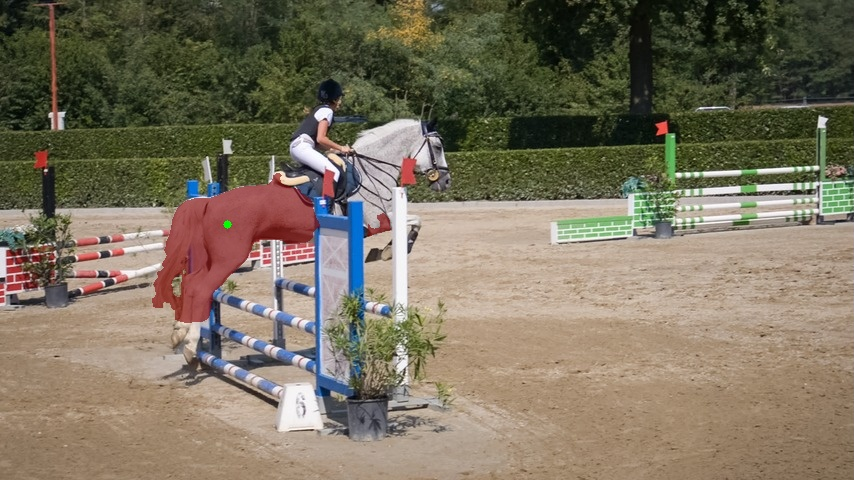}
    \subcaption*{1 click 42.9\%}
  \end{subfigure}
  \hspace{2mm}
  \begin{subfigure}[t]{0.22\linewidth}
    \includegraphics[width=1\linewidth]{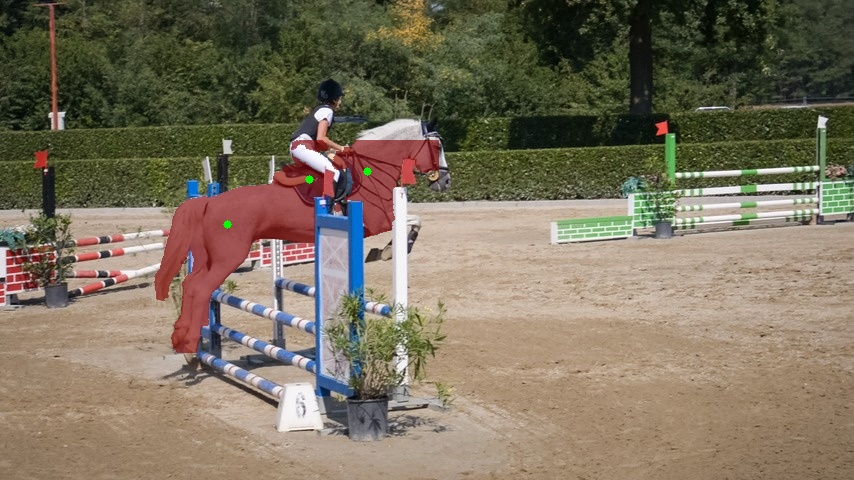}
    \subcaption*{3 clicks 69.6\%}
  \end{subfigure}
  \hspace{2mm}
  \begin{subfigure}[t]{0.22\linewidth}
    \includegraphics[width=1\linewidth]{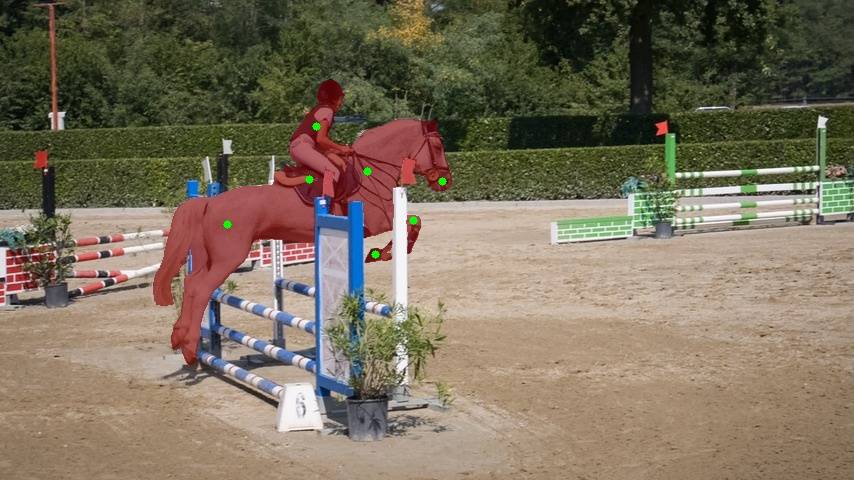}
    \subcaption*{7 clicks 90.3\%}
  \end{subfigure}
  \vspace{-2mm}
  \caption{More visualizations of the segmentation results from SBD \cite{hariharan2011sbd} (Row 1-4) and DAVIS \cite{perazzi2016davis} (Row 5-7).}
  \vspace{-3mm}
  \label{fig:segmentations more-sbd-davis}
\end{figure*}

\begin{figure*}[ht]
  \centering
  \begin{subfigure}[t]{0.42\linewidth}
    \includegraphics[width=1\linewidth]{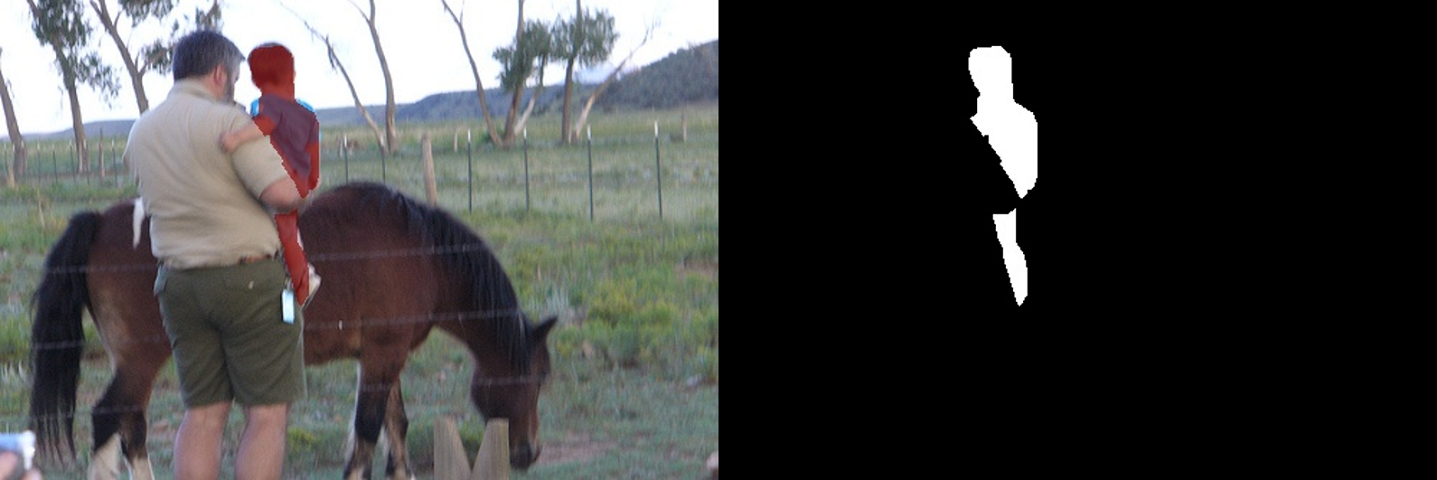}
    \subcaption*{GT}
  \end{subfigure}
  \hspace{2mm}
  \begin{subfigure}[t]{0.5\linewidth}
    \includegraphics[width=1\linewidth]{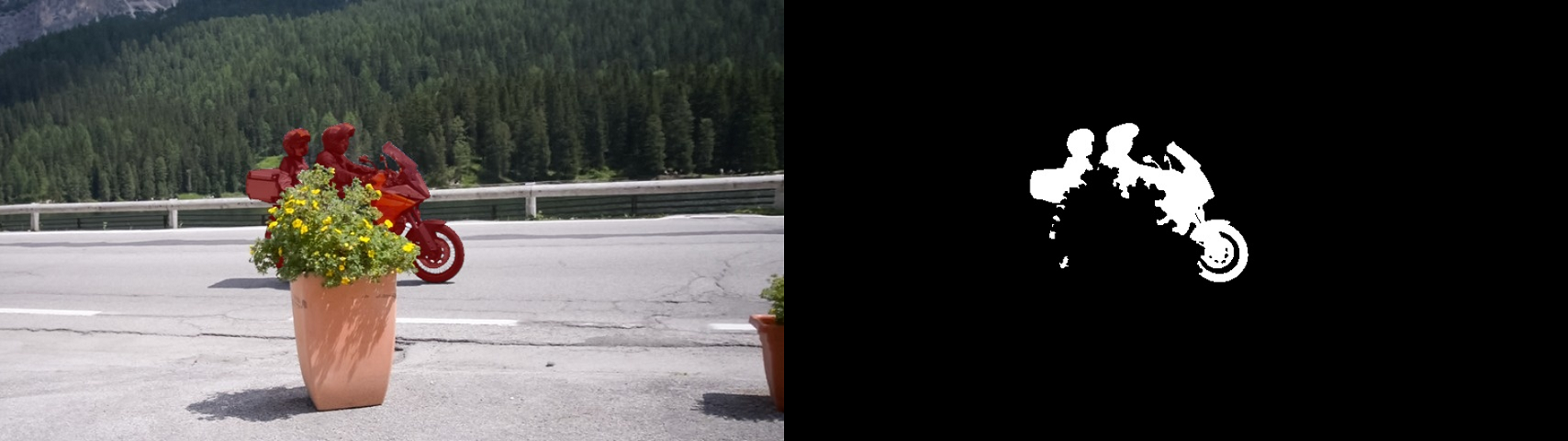}
    \subcaption*{GT}
  \end{subfigure}
  
  \begin{subfigure}[t]{0.42\linewidth}
    \includegraphics[width=1\linewidth]{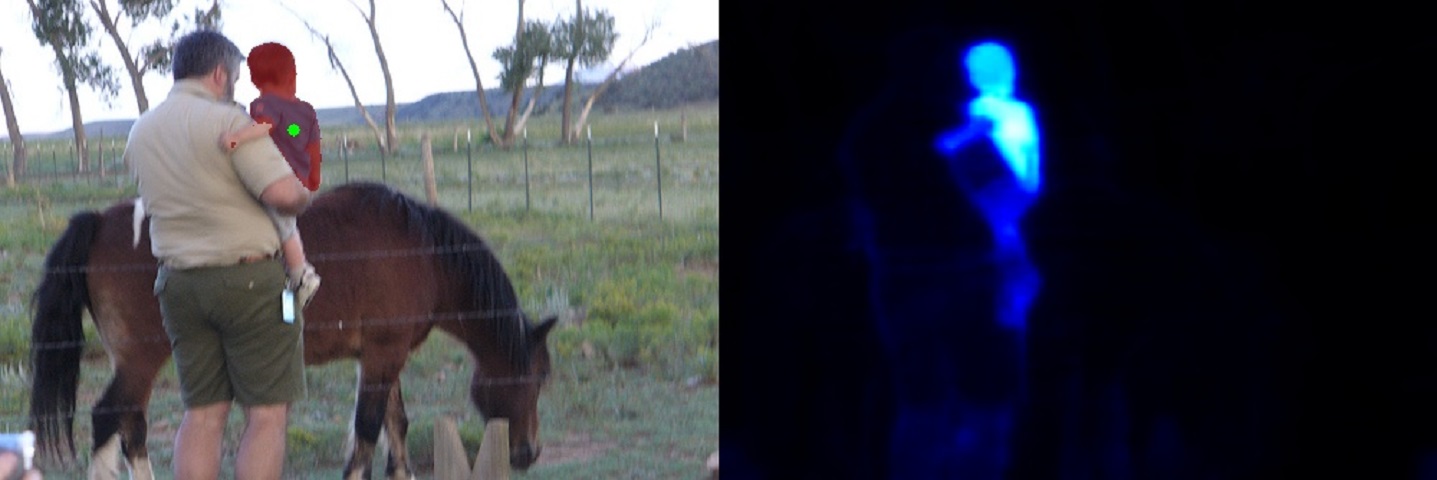}
    \subcaption*{1 click 66.2\%}
  \end{subfigure}
  \hspace{2mm}
  \begin{subfigure}[t]{0.5\linewidth}
    \includegraphics[width=1\linewidth]{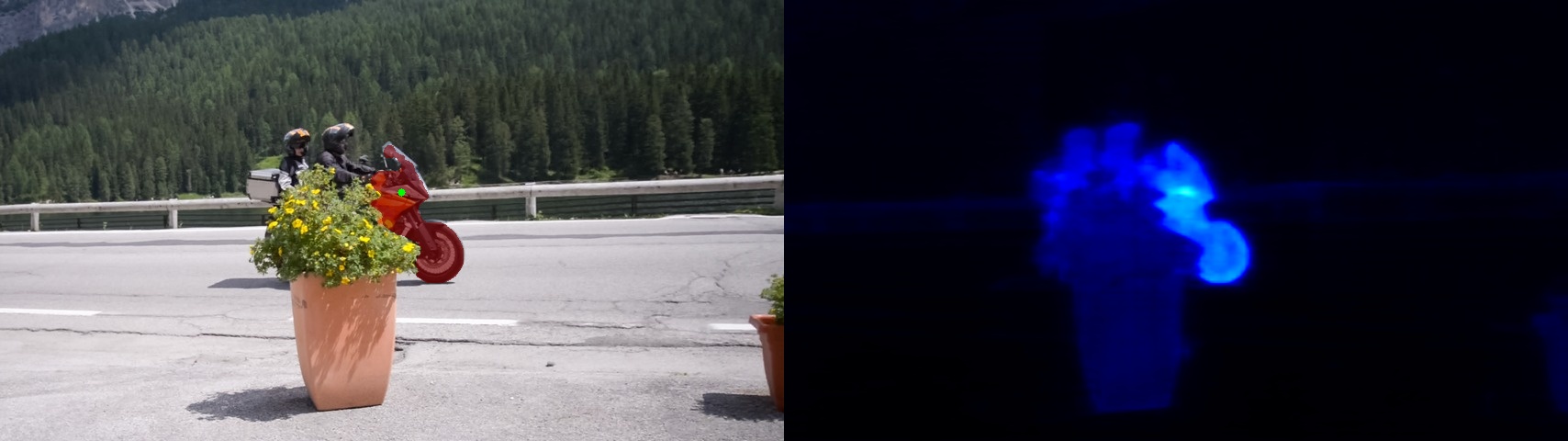}
    \subcaption*{1 click 45.2\%}
  \end{subfigure}

  \begin{subfigure}[t]{0.42\linewidth}
    \includegraphics[width=1\linewidth]{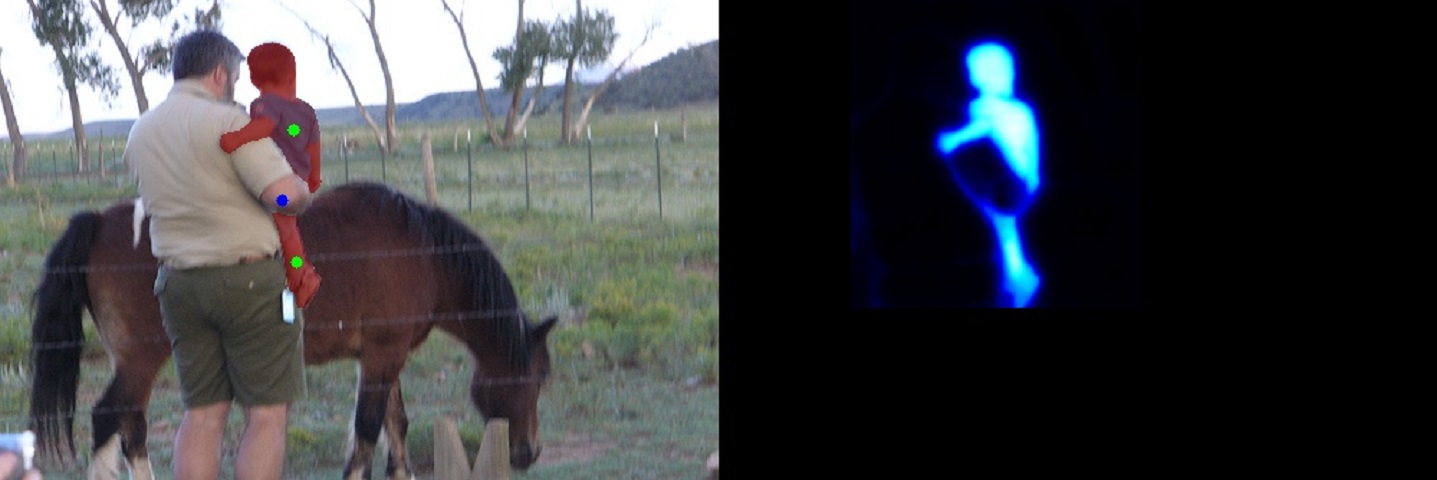}
    \subcaption*{3 clicks 73.9\%}
  \end{subfigure}
  \hspace{2mm}
  \begin{subfigure}[t]{0.5\linewidth}
    \includegraphics[width=1\linewidth]{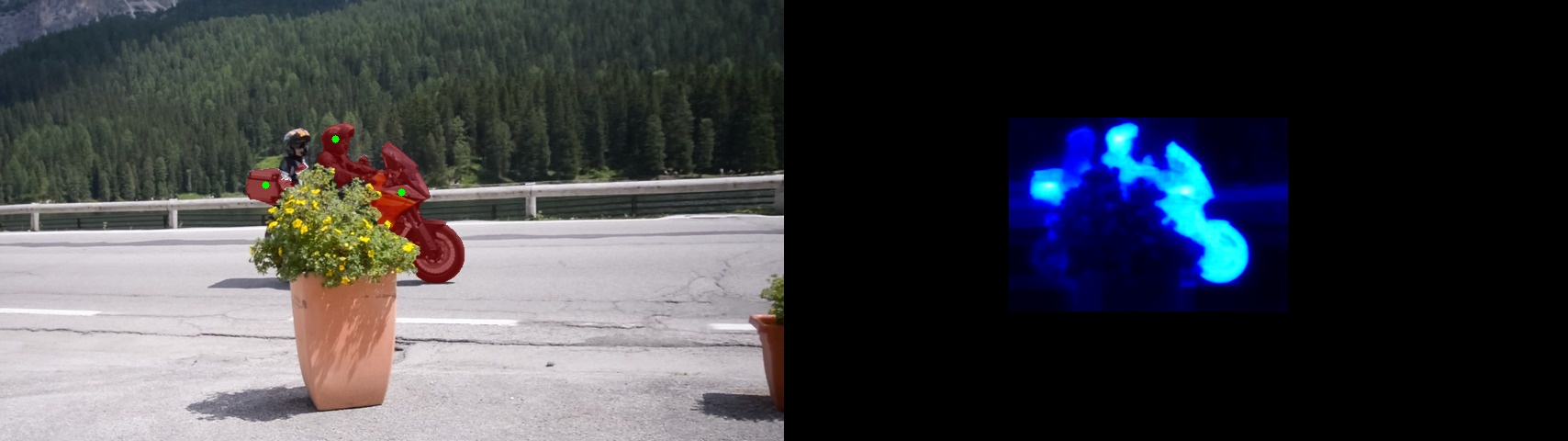}
    \subcaption*{3 clicks 74.8\%}
  \end{subfigure}

  \begin{subfigure}[t]{0.42\linewidth}
    \includegraphics[width=1\linewidth]{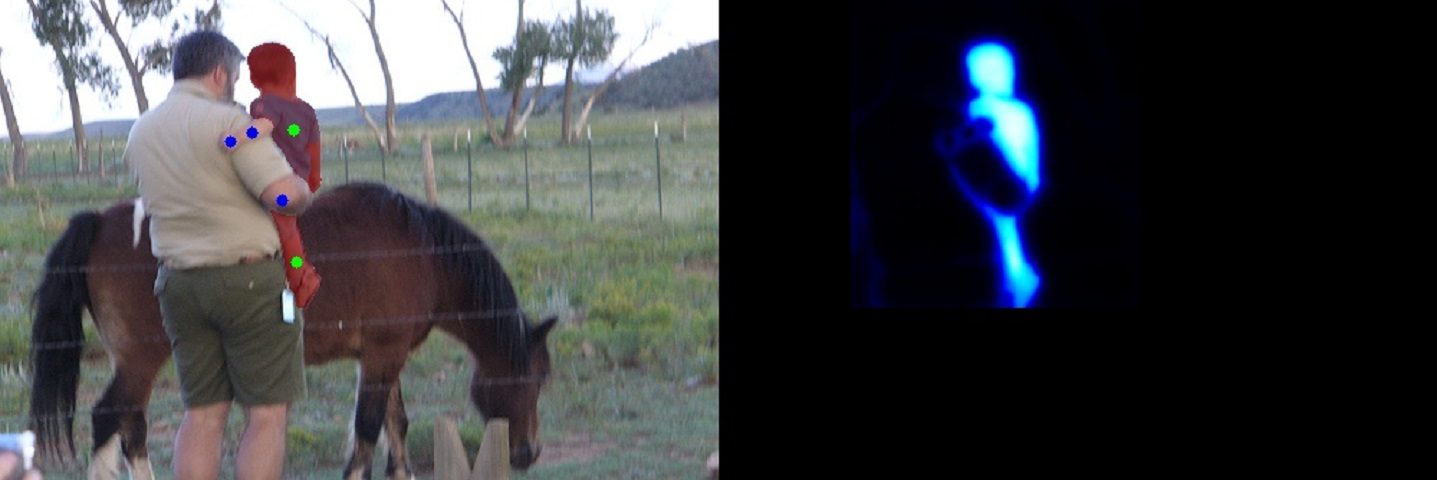}
    \subcaption*{5 clicks 77.2\%}
  \end{subfigure}
  \hspace{2mm}
  \begin{subfigure}[t]{0.5\linewidth}
    \includegraphics[width=1\linewidth]{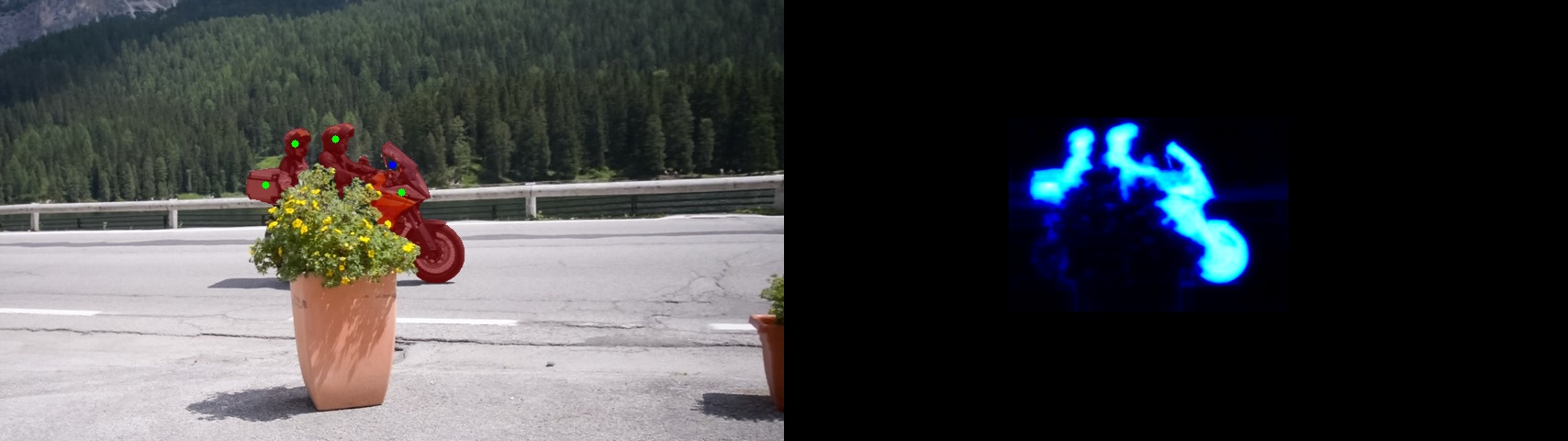}
    \subcaption*{5 clicks 85.7\%}
  \end{subfigure}

  \begin{subfigure}[t]{0.42\linewidth}
    \includegraphics[width=1\linewidth]{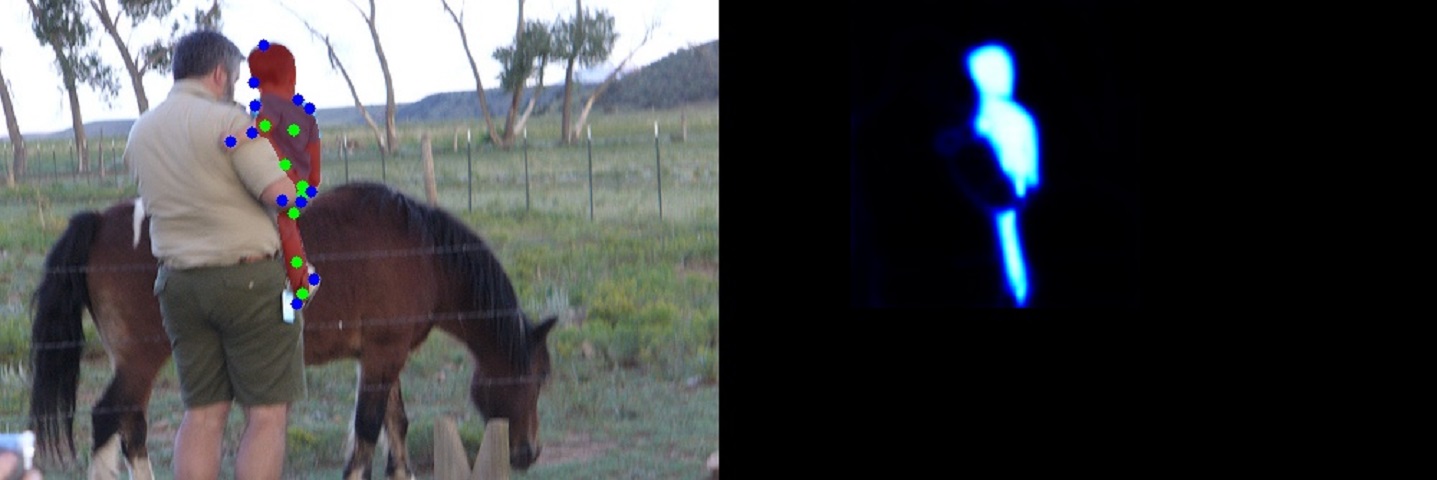}
    \subcaption*{20 clicks 90.5\%}
  \end{subfigure}
  \hspace{2mm}
  \begin{subfigure}[t]{0.5\linewidth}
    \includegraphics[width=1\linewidth]{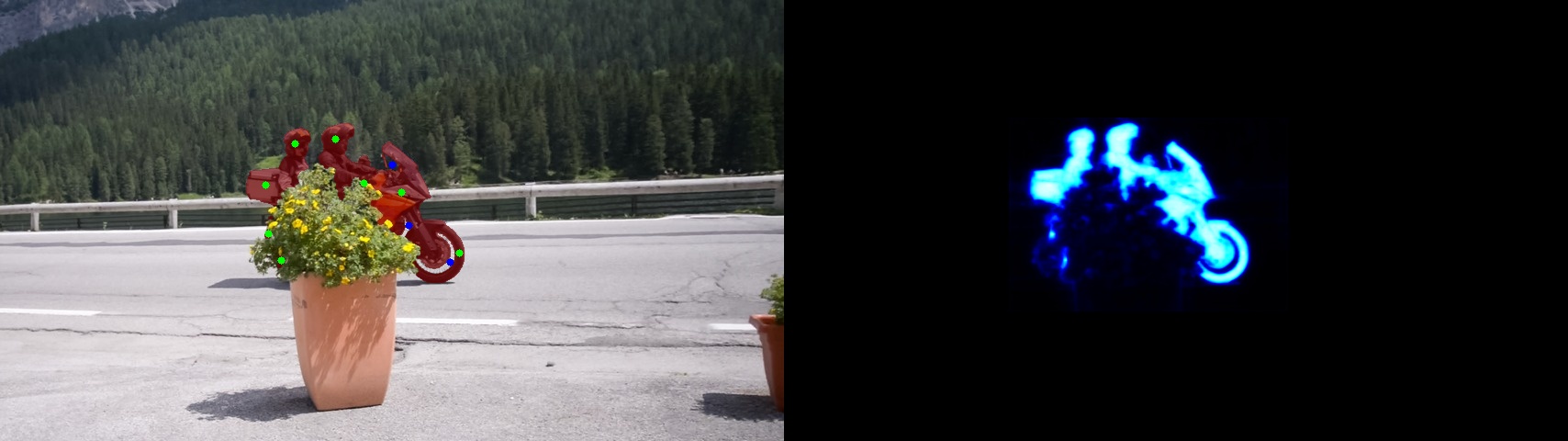}
    \subcaption*{11 clicks 90.0\%}
  \end{subfigure}
  \vspace{-2mm}
  \caption{Some of the challenging cases from SBD \cite{hariharan2011sbd} (left) and DAVIS \cite{perazzi2016davis} (right). Green and blue dots denote positive and negative clicks, respectively. The segmentation probability maps are displayed next to the images with overlaid masks.}
  \vspace{-3mm}
  \label{fig:challenging cases}
\end{figure*}

\begin{figure*}[ht]
  \centering
  \begin{subfigure}[t]{0.45\linewidth}
    \includegraphics[width=1\linewidth]{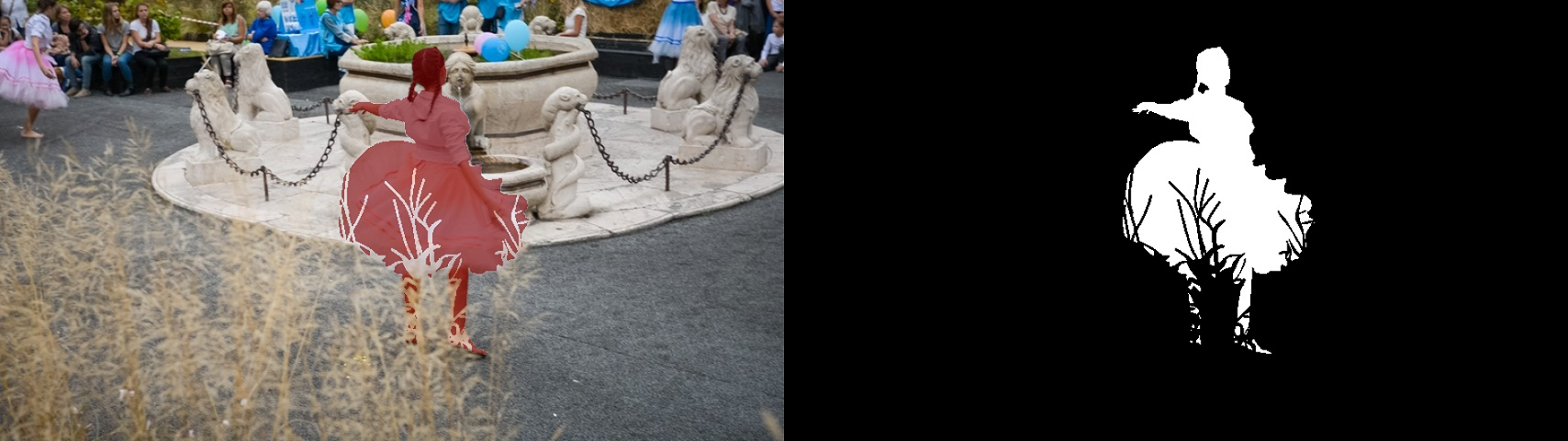}
    \subcaption*{GT}
  \end{subfigure}
  \hspace{2mm}
  \begin{subfigure}[t]{0.45\linewidth}
    \includegraphics[width=1\linewidth]{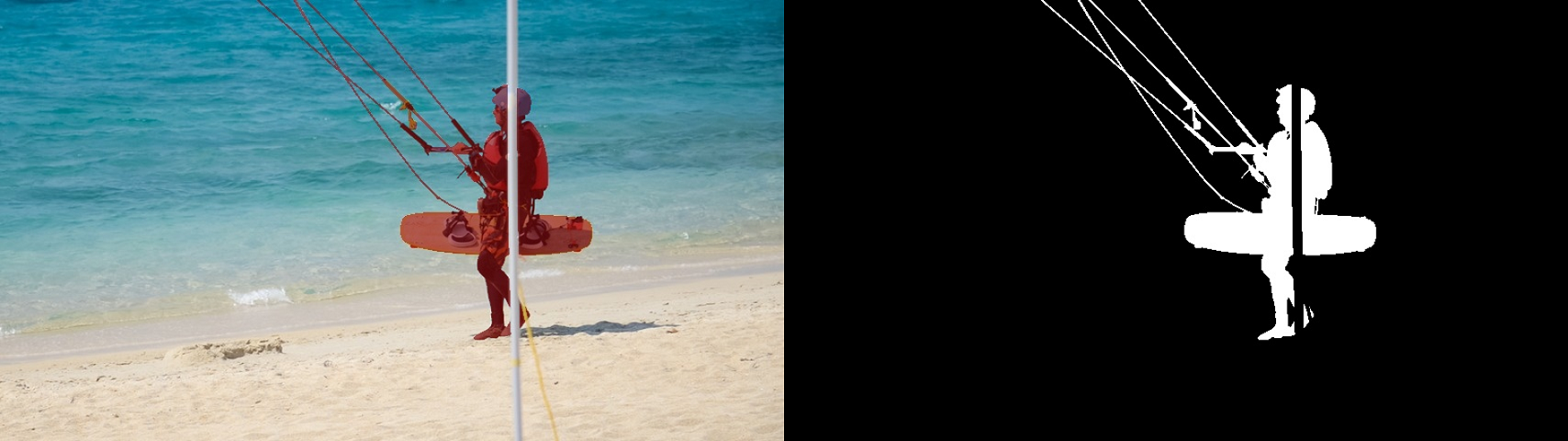}
    \subcaption*{GT}
  \end{subfigure}
  
  \begin{subfigure}[t]{0.45\linewidth}
    \includegraphics[width=1\linewidth]{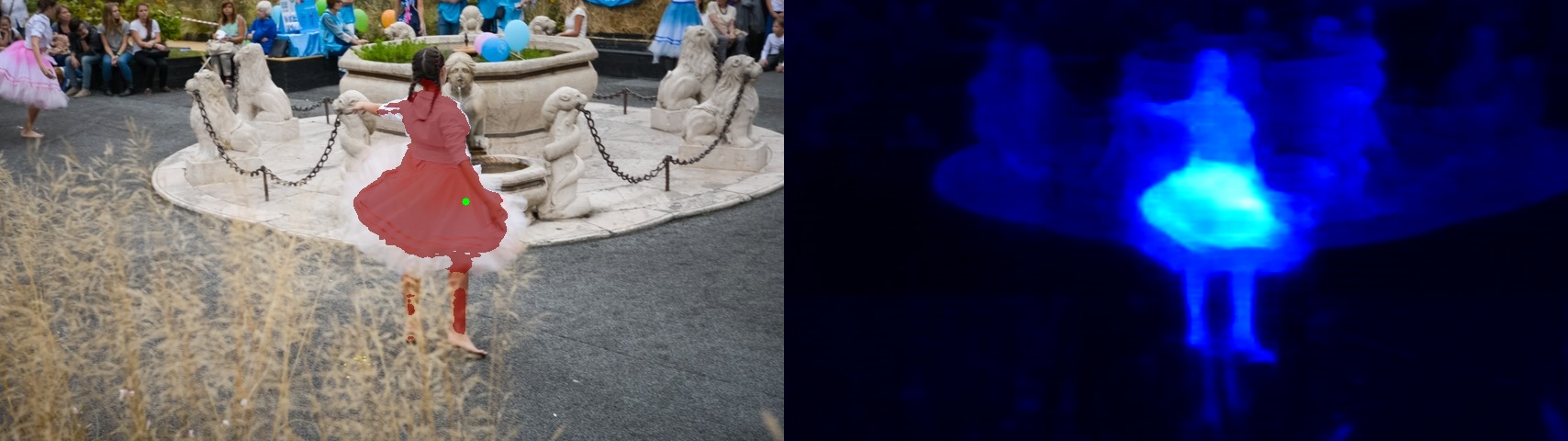}
    \subcaption*{1 click 62.5\%}
  \end{subfigure}
  \hspace{2mm}
  \begin{subfigure}[t]{0.45\linewidth}
    \includegraphics[width=1\linewidth]{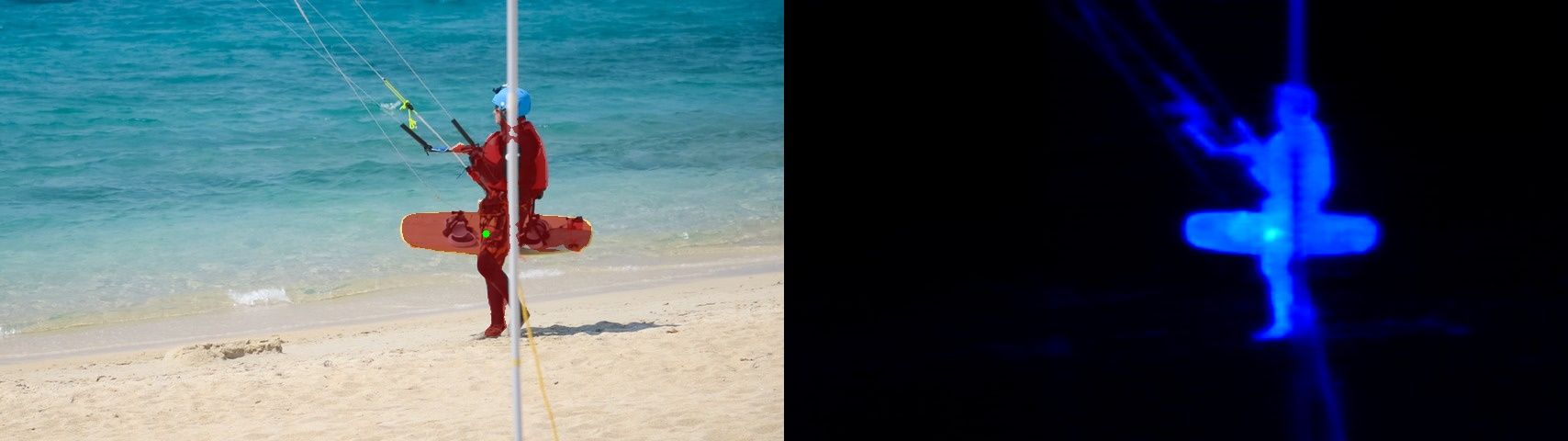}
    \subcaption*{1 click 70.6\%}
  \end{subfigure}

  \begin{subfigure}[t]{0.45\linewidth}
    \includegraphics[width=1\linewidth]{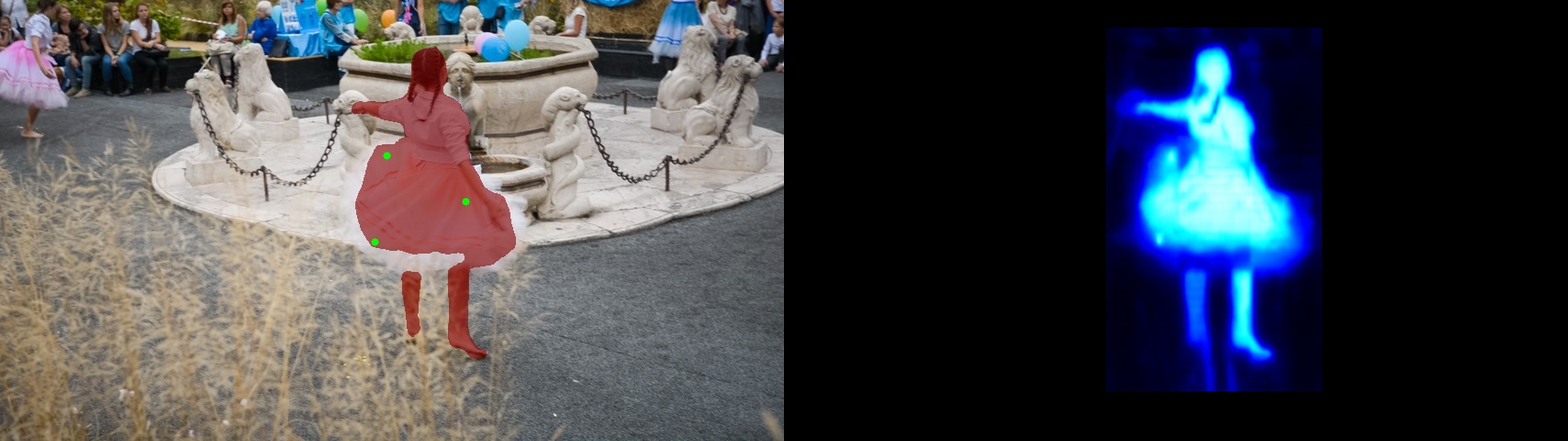}
    \subcaption*{3 clicks 73.4\%}
  \end{subfigure}
  \hspace{2mm}
  \begin{subfigure}[t]{0.45\linewidth}
    \includegraphics[width=1\linewidth]{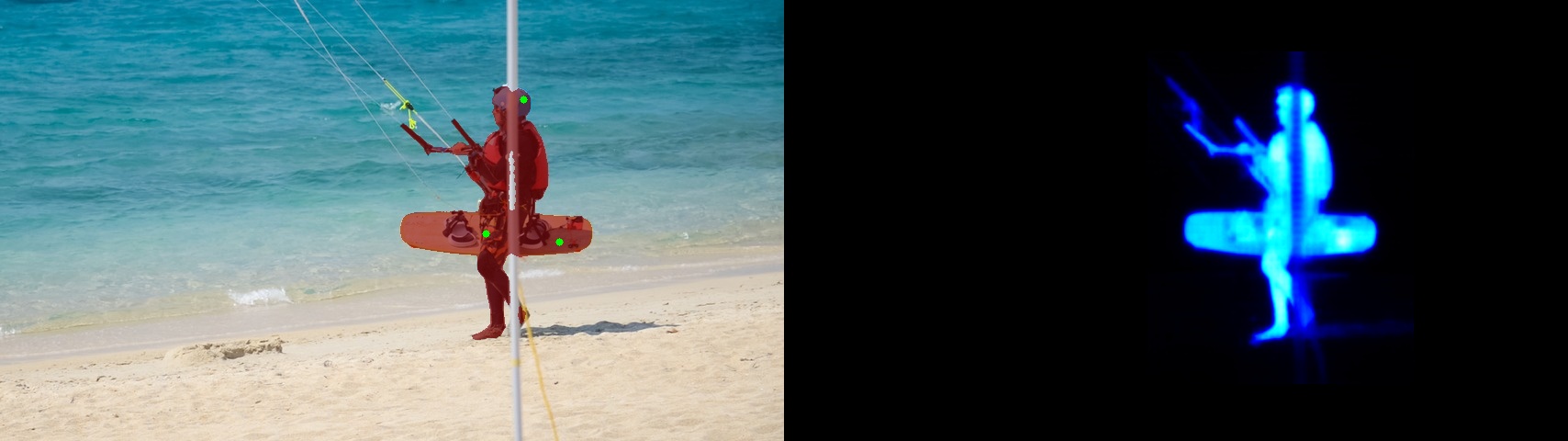}
    \subcaption*{3 clicks 78.7\%}
  \end{subfigure}

  \begin{subfigure}[t]{0.45\linewidth}
    \includegraphics[width=1\linewidth]{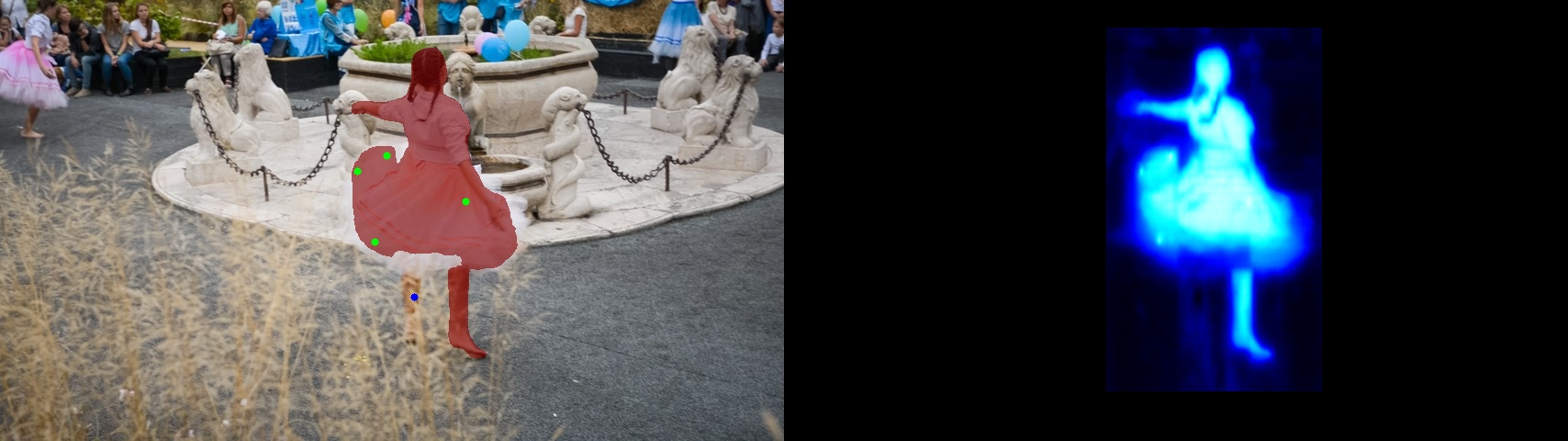}
    \subcaption*{5 clicks 78.1\%}
  \end{subfigure}
  \hspace{2mm}
  \begin{subfigure}[t]{0.45\linewidth}
    \includegraphics[width=1\linewidth]{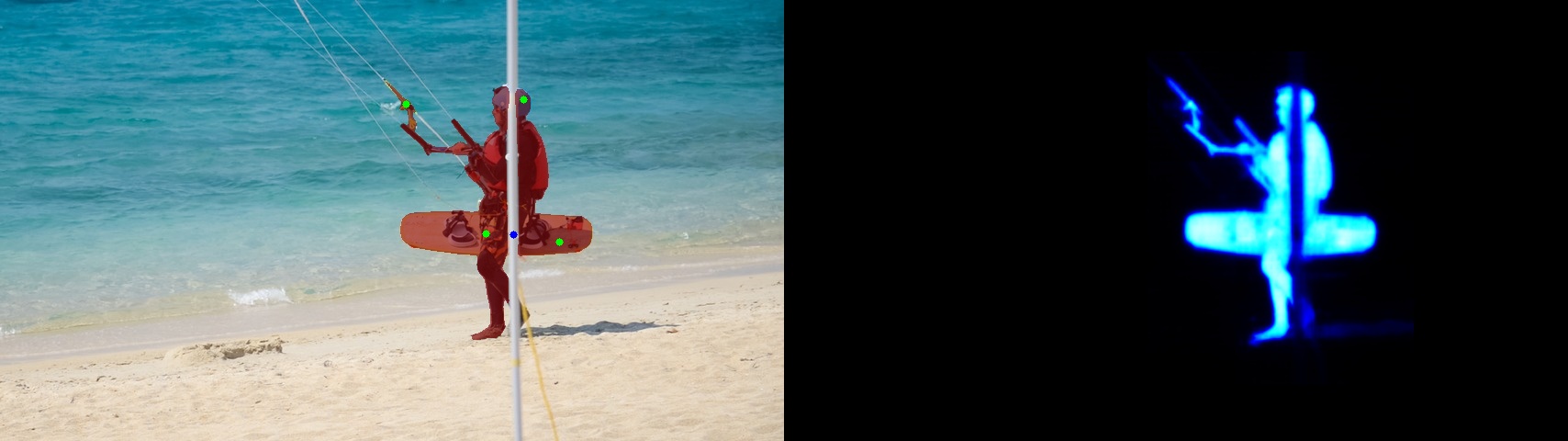}
    \subcaption*{5 clicks 83.2\%}
  \end{subfigure}

  \begin{subfigure}[t]{0.45\linewidth}
    \includegraphics[width=1\linewidth]{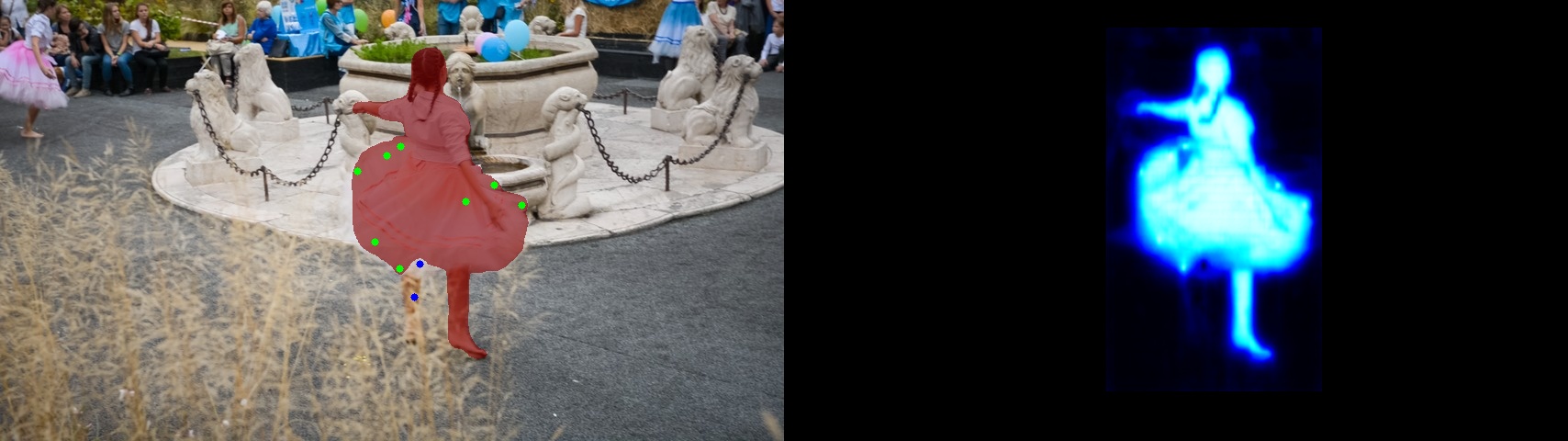}
    \subcaption*{10 clicks 81.1\%}
  \end{subfigure}
  \hspace{2mm}
  \begin{subfigure}[t]{0.45\linewidth}
    \includegraphics[width=1\linewidth]{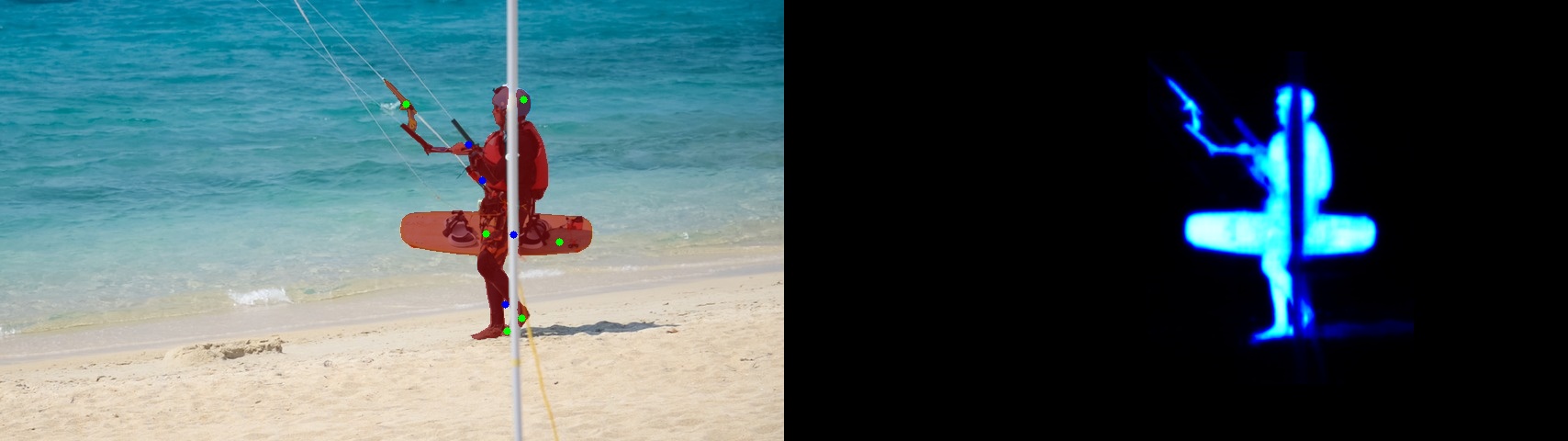}
    \subcaption*{10 clicks 82.2\%}
  \end{subfigure}

  \begin{subfigure}[t]{0.45\linewidth}
    \includegraphics[width=1\linewidth]{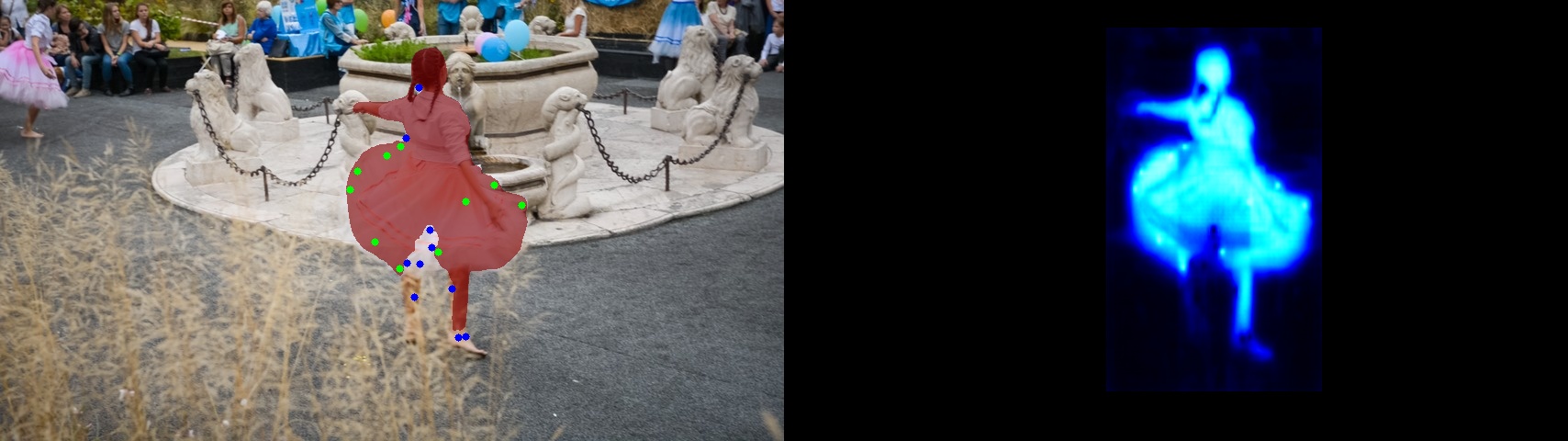}
    \subcaption*{20 clicks 84.6\%}
  \end{subfigure}
  \hspace{2mm}
  \begin{subfigure}[t]{0.45\linewidth}
    \includegraphics[width=1\linewidth]{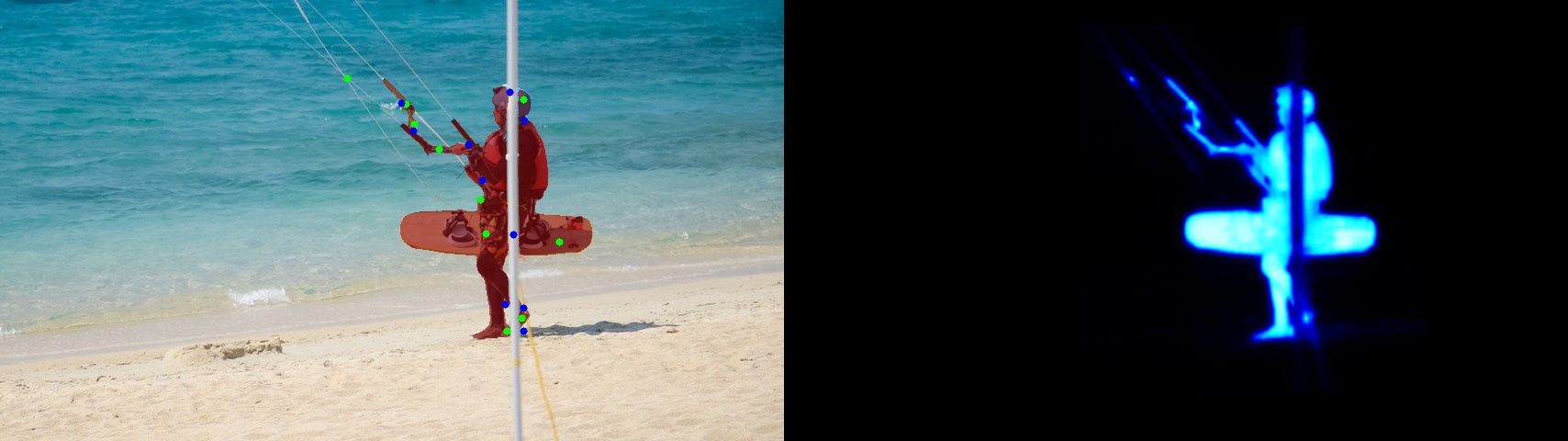}
    \subcaption*{20 clicks 83.1\%}
  \end{subfigure}
  \vspace{-2mm}
  \caption{Some of the bad cases from DAVIS \cite{perazzi2016davis}. The segmentation probability maps are displayed next to the images with overlaid masks.}
  \vspace{-3mm}
  \label{fig:bad cases}
\end{figure*}

We also provide more segmentation results of our iCMFormer on the 4 datasets. Fig.~\ref{fig:segmentations more-grabcut-berkeley} shows the common cases from GrabCut \cite{rother2004grabcut} and Berkeley \cite{martin2001berkely}, and Fig.~\ref{fig:segmentations more-sbd-davis} represents common cases from SBD \cite{hariharan2011sbd} and DAVIS \cite{perazzi2016davis}. As shown in Fig.~\ref{fig:challenging cases}, we display some challenging cases where it requires more than the average number of clicks to get the target IoU. We report the segmentation results in the middle stages until reaching 90\% IoU. However, there still exist some bad cases due to the limitations of our method, and Fig.~\ref{fig:bad cases} shows 2 examples from DAVIS.

\begin{table*}[t!]
  \caption{Comparison with previous models in term of number of failures (NoF) that cannot reach the target IoUs after 20 clicks, denoted as $\geq$20@85 and $\geq$20@90, respectively. The results are divided into 2 sections on the basis of the training datsets: SBD \cite{hariharan2011sbd} (represented as $\dagger$) and COCO \cite{lin2014coco} + LVIS \cite{gupta2019lvis} (represented as $\ddagger$). The best results are \textbf{bold}.}
  \centering
  \setlength\tabcolsep{4pt}
  \begin{tabular}{p{4cm}cccc}
  \toprule
  \multirow{2}*{Method} & \multicolumn{2}{c}{SBD} & \multicolumn{2}{c}{DAVIS}\\
 & $\geq$20@85 & $\geq$20@90 & $\geq$20@85 & $\geq$20@90\\
  \midrule
  BRS\cite{jang2019brs}$\dagger$ & - & - & - & 77\\
  f-BRS\cite{sofiiuk2020fbrs}$\dagger$ & - & 1466 & - & 78\\
  CDNet\cite{chen2021cdnet}$\dagger$ & - & - & 46 & 65\\
  FocusCut\cite{lin2022focuscut}$\dagger$ & - & - & - & 57\\
  Ours-ViT-B$\dagger$ & \textbf{236} & \textbf{693} & \textbf{30} & \textbf{53}\\
  Ours-Swin-B$\dagger$ & 242 & 698 & 36 & \textbf{53}\\
  \midrule\hline
  RITM-HRNet-18\cite{sofiiuk2022ritm}$\ddagger$ & - & - & 52 & 91\\
  FocalClick-HRNet-18\cite{chen2022focalclick}$\ddagger$ & - & - & 49 & 77\\
  FocalClick-SegF-B0-S2\cite{chen2022focalclick}$\ddagger$ & - & - & 50 & 86\\
  Ours-ViT-B$\ddagger$ & \textbf{225} & 695 & \textbf{20} & 49\\
  Ours-Swin-B$\ddagger$ & 237 & \textbf{667} & \textbf{20} & \textbf{48}\\
  \bottomrule
  \end{tabular}
  \label{tab:failure 2 datasets}
\end{table*}
\begin{table*}[t!]
  \caption{Ablation study for the number of proposed cross-modality blocks on GrabCut \cite{rother2004grabcut}, Berkeley \cite{martin2001berkely}, SBD \cite{hariharan2011sbd} and DAVIS \cite{perazzi2016davis} datasets. NoC85 and NoC90 denote the average numbers of clicks to reach a target IoU. All the models are trained on SBD. The best results are \textbf{bold} while the second best are \underline{underlined}.}
  \centering
  \setlength\tabcolsep{4pt}
  \begin{tabular}{p{2cm}cccccccccc}
  \toprule
  \multirow{2}*{Method} & \multirow{2}*{Layer} & \multirow{2}*{Params/M} & \multicolumn{2}{c}{GrabCut} & \multicolumn{2}{c}{Berkeley} & \multicolumn{2}{c}{SBD} & \multicolumn{2}{c}{DAVIS}\\
 & & & NoC85 &  NoC90 & NoC85 &  NoC90 & NoC85 &  NoC90 & NoC85 &  NoC90\\
  \midrule
  Ours-ViT-B & 1 & 105.90 & 1.46 & 1.68 & 1.50 & 2.56 & \textbf{3.28} & \textbf{5.25} & 4.20 & 5.60\\
  Ours-ViT-B & 2 & 115.36 & \underline{1.44} & \underline{1.52} & \underline{1.46} & 2.55 & \underline{3.32} & 5.31 & \underline{4.09} & 5.62\\
  Ours-ViT-B & 3 & 124.81 & \textbf{1.36} & \textbf{1.42} & \textbf{1.42} & \underline{2.52} & 3.33 & 5.31 & \textbf{4.05} & \underline{5.58}\\
  Ours-ViT-B & 6 & 153.16 & 1.52 & 1.58 & 1.47 & 2.54 & 3.37 & 5.36 & 4.17 & 5.75\\
  Ours-ViT-B & 8 & 172.07 & 1.54 & 1.66 & 1.59 & \textbf{2.45} & \underline{3.32} & \underline{5.30} & 4.10 & \underline{5.54}\\
  \midrule
  Ours-Swin-B & 1 & 91.64 & 1.48 & \underline{1.56} & 1.56 & 2.57 & 3.31 & 5.41 & 4.38 & 6.07\\
  Ours-Swin-B & 2 & 95.84 & \underline{1.42} & 1.62 & 1.56 & 2.58 & 3.28 & \underline{5.25} & \textbf{4.18} & 5.70\\
  Ours-Swin-B & 4 & 104.25 & 1.46 & \textbf{1.50} & \textbf{1.52} & \textbf{2.32} & \textbf{3.21} & \textbf{5.16} & \underline{4.25} & \textbf{5.55}\\
  Ours-Swin-B & 6 & 112.66 & 1.46 & 1.62 & \underline{1.55} & 2.64 & \underline{3.24} & 5.29 & 4.34 & 5.68\\
  Ours-Swin-B & 8 & 121.06 & \textbf{1.40} & 1.62 & \underline{1.55} & \underline{2.50} & 3.28 & 5.34 & \underline{4.25} & \underline{5.67}\\
  \bottomrule
  \end{tabular}
  \label{tab:ablation study3}
\end{table*}

\newpage

\newpage

{\small
\bibliographystyle{ieee_fullname}
\bibliography{egbib}

\begin{thebibliography}{10}\itemsep=-1pt

\bibitem{acuna2018poligonrnn+}
David Acuna, Huan Ling, Amlan Kar, and Sanja Fidler.
\newblock Efficient interactive annotation of segmentation datasets with
  polygon-rnn++.
\newblock In {\em Proceedings of the IEEE Conference on Computer Vision and
  Pattern Recognition}, pages 859--868, 2018.

\bibitem{ahmed2014objectselection}
Ejaz Ahmed, Scott Cohen, and Brian Price.
\newblock Semantic object selection.
\newblock In {\em Proceedings of the IEEE Conference on Computer Vision and
  Pattern Recognition}, pages 3150--3157, 2014.

\bibitem{antol2015vqa}
Stanislaw Antol, Aishwarya Agrawal, Jiasen Lu, Margaret Mitchell, Dhruv Batra,
  C~Lawrence Zitnick, and Devi Parikh.
\newblock Vqa: Visual question answering.
\newblock In {\em Proceedings of the IEEE International Conference on Computer
  Vision}, pages 2425--2433, 2015.

\bibitem{asad2022scrribles}
Muhammad Asad, Lucas Fidon, and Tom Vercauteren.
\newblock Econet: Efficient convolutional online likelihood network for
  scribble-based interactive segmentation.
\newblock In {\em International Conference on Medical Imaging with Deep
  Learning}, pages 35--47, 2022.

\bibitem{chen2021cdnet}
Xi Chen, Zhiyan Zhao, Feiwu Yu, Yilei Zhang, and Manni Duan.
\newblock Conditional diffusion for interactive segmentation.
\newblock In {\em Proceedings of the IEEE International Conference on Computer
  Vision}, pages 7345--7354, 2021.

\bibitem{chen2022focalclick}
Xi Chen, Zhiyan Zhao, Yilei Zhang, Manni Duan, Donglian Qi, and Hengshuang
  Zhao.
\newblock Focalclick: towards practical interactive image segmentation.
\newblock In {\em Proceedings of the IEEE Conference on Computer Vision and
  Pattern Recognition}, pages 1300--1309, 2022.

\bibitem{chu2021twins}
Xiangxiang Chu, Zhi Tian, Yuqing Wang, Bo Zhang, Haibing Ren, Xiaolin Wei,
  Huaxia Xia, and Chunhua Shen.
\newblock Twins: Revisiting the design of spatial attention in vision
  transformers.
\newblock {\em Advances in Neural Information Processing Systems}, pages
  9355--9366, 2021.

\bibitem{dosovitskiy2020vit}
Alexey Dosovitskiy, Lucas Beyer, Alexander Kolesnikov, Dirk Weissenborn,
  Xiaohua Zhai, Thomas Unterthiner, Mostafa Dehghani, Matthias Minderer, Georg
  Heigold, Sylvain Gelly, et~al.
\newblock An image is worth 16x16 words: Transformers for image recognition at
  scale.
\newblock {\em arXiv preprint arXiv:2010.11929}, 2020.

\bibitem{du2023emc}
Fei Du, Jianlong Yuan, Zhibin Wang, and Fan Wang.
\newblock Efficient mask correction for click-based interactive image
  segmentation.
\newblock In {\em Proceedings of the IEEE/CVF Conference on Computer Vision and
  Pattern Recognition}, pages 22773--22782, 2023.

\bibitem{everingham2009pascal}
Mark Everingham, Luc Van~Gool, Christopher~KI Williams, John Winn, and Andrew
  Zisserman.
\newblock The pascal visual object classes (voc) challenge.
\newblock {\em International Journal of Computer Vision}, 88:303--308, 2009.

\bibitem{grady2006random}
Leo Grady.
\newblock Random walks for image segmentation.
\newblock {\em IEEE Transactions on Pattern Analysis and Machine Intelligence},
  28(11):1768--1783, 2006.

\bibitem{gupta2019lvis}
Agrim Gupta, Piotr Dollar, and Ross Girshick.
\newblock Lvis: A dataset for large vocabulary instance segmentation.
\newblock In {\em Proceedings of the IEEE Conference on Computer Vision and
  Pattern Recognition}, pages 5356--5364, 2019.

\bibitem{hariharan2011sbd}
Bharath Hariharan, Pablo Arbel{\'a}ez, Lubomir Bourdev, Subhransu Maji, and
  Jitendra Malik.
\newblock Semantic contours from inverse detectors.
\newblock In {\em Proceedings of the IEEE International Conference on Computer
  Vision}, pages 991--998, 2011.

\bibitem{he2016resnet}
Kaiming He, Xiangyu Zhang, Shaoqing Ren, and Jian Sun.
\newblock Deep residual learning for image recognition.
\newblock In {\em Proceedings of the IEEE Conference on Computer Vision and
  Pattern Recognition}, pages 770--778, 2016.

\bibitem{jang2019brs}
Won-Dong Jang and Chang-Su Kim.
\newblock Interactive image segmentation via backpropagating refinement scheme.
\newblock In {\em Proceedings of the IEEE Conference on Computer Vision and
  Pattern Recognition}, pages 5297--5306, 2019.

\bibitem{lee2020imageedit}
Cheng-Han Lee, Ziwei Liu, Lingyun Wu, and Ping Luo.
\newblock Maskgan: Towards diverse and interactive facial image manipulation.
\newblock In {\em Proceedings of the IEEE Conference on Computer Vision and
  Pattern Recognition}, pages 5549--5558, 2020.

\bibitem{li2022vitdet}
Yanghao Li, Hanzi Mao, Ross Girshick, and Kaiming He.
\newblock Exploring plain vision transformer backbones for object detection.
\newblock In {\em European Conference on Computer Vision}, pages 280--296.
  Springer, 2022.

\bibitem{li2004lazy}
Yin Li, Jian Sun, Chi-Keung Tang, and Heung-Yeung Shum.
\newblock Lazy snapping.
\newblock {\em ACM Transactions on Graphics (ToG)}, 23(3):303--308, 2004.

\bibitem{li2018latent}
Zhuwen Li, Qifeng Chen, and Vladlen Koltun.
\newblock Interactive image segmentation with latent diversity.
\newblock In {\em Proceedings of the IEEE Conference on Computer Vision and
  Pattern Recognition}, pages 577--585, 2018.

\bibitem{liew2017ris-net}
JunHao Liew, Yunchao Wei, Wei Xiong, Sim-Heng Ong, and Jiashi Feng.
\newblock Regional interactive image segmentation networks.
\newblock In {\em Proceedings of the IEEE International Conference on Computer
  Vision}, pages 2746--2754, 2017.

\bibitem{lin2014coco}
Tsung-Yi Lin, Michael Maire, Serge Belongie, James Hays, Pietro Perona, Deva
  Ramanan, Piotr Doll{\'a}r, and C~Lawrence Zitnick.
\newblock Microsoft coco: Common objects in context.
\newblock In {\em European Conference on Computer Vision}, pages 740--755.
  Springer, 2014.

\bibitem{lin2022focuscut}
Zheng Lin, Zheng-Peng Duan, Zhao Zhang, Chun-Le Guo, and Ming-Ming Cheng.
\newblock Focuscut: Diving into a focus view in interactive segmentation.
\newblock In {\em Proceedings of the IEEE Conference on Computer Vision and
  Pattern Recognition}, pages 2637--2646, 2022.

\bibitem{lin2020fca}
Zheng Lin, Zhao Zhang, Lin-Zhuo Chen, Ming-Ming Cheng, and Shao-Ping Lu.
\newblock Interactive image segmentation with first click attention.
\newblock In {\em Proceedings of the IEEE Conference on Computer Vision and
  Pattern Recognition}, pages 13339--13348, 2020.

\bibitem{liu2022pseudoclick}
Qin Liu, Meng Zheng, Benjamin Planche, Srikrishna Karanam, Terrence Chen, Marc
  Niethammer, and Ziyan Wu.
\newblock Pseudoclick: Interactive image segmentation with click imitation.
\newblock In {\em European Conference on Computer Vision}, pages 728--745.
  Springer, 2022.

\bibitem{liu2018instanceseg}
Shu Liu, Lu Qi, Haifang Qin, Jianping Shi, and Jiaya Jia.
\newblock Path aggregation network for instance segmentation.
\newblock In {\em Proceedings of the IEEE Conference on Computer Vision and
  Pattern Recognition}, pages 8759--8768, 2018.

\bibitem{liu2021swin}
Ze Liu, Yutong Lin, Yue Cao, Han Hu, Yixuan Wei, Zheng Zhang, Stephen Lin, and
  Baining Guo.
\newblock Swin transformer: Hierarchical vision transformer using shifted
  windows.
\newblock In {\em Proceedings of the IEEE International Conference on Computer
  Vision}, pages 10012--10022, 2021.

\bibitem{long2015fcn}
Jonathan Long, Evan Shelhamer, and Trevor Darrell.
\newblock Fully convolutional networks for semantic segmentation.
\newblock In {\em Proceedings of the IEEE Conference on Computer Vision and
  Pattern Recognition}, pages 3431--3440, 2015.

\bibitem{mahadevan2018itis}
Sabarinath Mahadevan, Paul Voigtlaender, and Bastian Leibe.
\newblock Iteratively trained interactive segmentation.
\newblock {\em arXiv preprint arXiv:1805.04398}, 2018.

\bibitem{majumder2019content}
Soumajit Majumder and Angela Yao.
\newblock Content-aware multi-level guidance for interactive instance
  segmentation.
\newblock In {\em Proceedings of the IEEE Conference on Computer Vision and
  Pattern Recognition}, pages 11602--11611, 2019.

\bibitem{maninis2018dextr}
Kevis-Kokitsi Maninis, Sergi Caelles, Jordi Pont-Tuset, and Luc Van~Gool.
\newblock Deep extreme cut: From extreme points to object segmentation.
\newblock In {\em Proceedings of the IEEE Conference on Computer Vision and
  Pattern Recognition}, pages 616--625, 2018.

\bibitem{martin2001berkely}
David Martin, Charless Fowlkes, Doron Tal, and Jitendra Malik.
\newblock A database of human segmented natural images and its application to
  evaluating segmentation algorithms and measuring ecological statistics.
\newblock In {\em Proceedings of the IEEE International Conference on Computer
  Vision}, pages 416--423, 2001.

\bibitem{ngiam2011multimodal}
Jiquan Ngiam, Aditya Khosla, Mingyu Kim, Juhan Nam, Honglak Lee, and Andrew~Y
  Ng.
\newblock Multimodal deep learning.
\newblock In {\em Proceedings of the International Conference on Machine
  Learning}, pages 689--696, 2011.

\bibitem{perazzi2016davis}
Federico Perazzi, Jordi Pont-Tuset, Brian McWilliams, Luc Van~Gool, Markus
  Gross, and Alexander Sorkine-Hornung.
\newblock A benchmark dataset and evaluation methodology for video object
  segmentation.
\newblock In {\em Proceedings of the IEEE Conference on Computer Vision and
  Pattern Recognition}, pages 724--732, 2016.

\bibitem{qian2020lidar}
Rui Qian, Divyansh Garg, Yan Wang, Yurong You, Serge Belongie, Bharath
  Hariharan, Mark Campbell, Kilian~Q Weinberger, and Wei-Lun Chao.
\newblock End-to-end pseudo-lidar for image-based 3d object detection.
\newblock In {\em Proceedings of the IEEE Conference on Computer Vision and
  Pattern Recognition}, pages 5881--5890, 2020.

\bibitem{rother2004grabcut}
Carsten Rother, Vladimir Kolmogorov, and Andrew Blake.
\newblock "grabcut" interactive foreground extraction using iterated graph
  cuts.
\newblock {\em ACM Transactions on Graphics (TOG)}, 23(3):309--314, 2004.

\bibitem{shen2017medical}
Dinggang Shen, Guorong Wu, and Heung-Il Suk.
\newblock Deep learning in medical image analysis.
\newblock {\em Annual Review of Biomedical Engineering}, 19:221--248, 2017.

\bibitem{siam2018autodrive}
Mennatullah Siam, Mostafa Gamal, Moemen Abdel-Razek, Senthil Yogamani, Martin
  Jagersand, and Hong Zhang.
\newblock A comparative study of real-time semantic segmentation for autonomous
  driving.
\newblock In {\em Proceedings of the IEEE Conference on Computer Vision and
  Pattern Recognition Workshops}, pages 587--597, 2018.

\bibitem{simonyan2014vgg}
Karen Simonyan and Andrew Zisserman.
\newblock Very deep convolutional networks for large-scale image recognition.
\newblock {\em arXiv preprint arXiv:1409.1556}, 2014.

\bibitem{sofiiuk2019nfl}
Konstantin Sofiiuk, Olga Barinova, and Anton Konushin.
\newblock Adaptis: Adaptive instance selection network.
\newblock In {\em Proceedings of the IEEE International Conference on Computer
  Vision}, pages 7355--7363, 2019.

\bibitem{sofiiuk2020fbrs}
Konstantin Sofiiuk, Ilia Petrov, Olga Barinova, and Anton Konushin.
\newblock f-brs: Rethinking backpropagating refinement for interactive
  segmentation.
\newblock In {\em Proceedings of the IEEE Conference on Computer Vision and
  Pattern Recognition}, pages 8623--8632, 2020.

\bibitem{sofiiuk2022ritm}
Konstantin Sofiiuk, Ilya~A Petrov, and Anton Konushin.
\newblock Reviving iterative training with mask guidance for interactive
  segmentation.
\newblock In {\em IEEE International Conference on Image Processing (ICIP)},
  pages 3141--3145, 2022.

\bibitem{vaswani2017transformer}
Ashish Vaswani, Noam Shazeer, Niki Parmar, Jakob Uszkoreit, Llion Jones,
  Aidan~N Gomez, {\L}ukasz Kaiser, and Illia Polosukhin.
\newblock Attention is all you need.
\newblock {\em Advances in Neural Information Processing Systems}, 30, 2017.

\bibitem{wang2021pyramid-t}
Wenhai Wang, Enze Xie, Xiang Li, Deng-Ping Fan, Kaitao Song, Ding Liang, Tong
  Lu, Ping Luo, and Ling Shao.
\newblock Pyramid vision transformer: A versatile backbone for dense prediction
  without convolutions.
\newblock In {\em Proceedings of the IEEE International Conference on Computer
  Vision}, pages 568--578, 2021.

\bibitem{wei2023fcf}
Qiaoqiao Wei, Hui Zhang, and Jun-Hai Yong.
\newblock Focused and collaborative feedback integration for interactive image
  segmentation.
\newblock In {\em Proceedings of the IEEE/CVF Conference on Computer Vision and
  Pattern Recognition}, pages 18643--18652, 2023.

\bibitem{xie2021segformer}
Enze Xie, Wenhai Wang, Zhiding Yu, Anima Anandkumar, Jose~M Alvarez, and Ping
  Luo.
\newblock Segformer: Simple and efficient design for semantic segmentation with
  transformers.
\newblock {\em Advances in Neural Information Processing Systems}, pages
  12077--12090, 2021.

\bibitem{xu2017deepgrabcut}
Ning Xu, Brian Price, Scott Cohen, Jimei Yang, and Thomas Huang.
\newblock Deep grabcut for object selection.
\newblock {\em arXiv preprint arXiv:1707.00243}, 2017.

\bibitem{xu2016dios}
Ning Xu, Brian Price, Scott Cohen, Jimei Yang, and Thomas~S Huang.
\newblock Deep interactive object selection.
\newblock In {\em Proceedings of the IEEE Conference on Computer Vision and
  Pattern Recognition}, pages 373--381, 2016.

\bibitem{yu2019mcan}
Zhou Yu, Jun Yu, Yuhao Cui, Dacheng Tao, and Qi Tian.
\newblock Deep modular co-attention networks for visual question answering.
\newblock In {\em Proceedings of the IEEE Conference on Computer Vision and
  Pattern Recognition}, pages 6281--6290, 2019.

\bibitem{yuan2021hrformer}
Yuhui Yuan, Rao Fu, Lang Huang, Weihong Lin, Chao Zhang, Xilin Chen, and
  Jingdong Wang.
\newblock Hrformer: High-resolution vision transformer for dense predict.
\newblock {\em Advances in Neural Information Processing Systems}, pages
  7281--7293, 2021.

\bibitem{yuan2022cross-ic}
Zhihao Yuan, Xu Yan, Yinghong Liao, Yao Guo, Guanbin Li, Shuguang Cui, and Zhen
  Li.
\newblock X-trans2cap: Cross-modal knowledge transfer using transformer for 3d
  dense captioning.
\newblock In {\em Proceedings of the IEEE Conference on Computer Vision and
  Pattern Recognition}, pages 8563--8573, 2022.

\bibitem{zhou2020unifiedvqa}
Luowei Zhou, Hamid Palangi, Lei Zhang, Houdong Hu, Jason Corso, and Jianfeng
  Gao.
\newblock Unified vision-language pre-training for image captioning and vqa.
\newblock In {\em Proceedings of the AAAI Conference on Artificial
  Intelligence}, volume~34, pages 13041--13049, 2020.

\bibitem{zhou2023gaussion}
Minghao Zhou, Hong Wang, Qian Zhao, Yuexiang Li, Yawen Huang, Deyu Meng, and
  Yefeng Zheng.
\newblock Interactive segmentation as gaussion process classification.
\newblock In {\em Proceedings of the IEEE/CVF Conference on Computer Vision and
  Pattern Recognition}, pages 19488--19497, 2023.

\end{thebibliography}
}

\end{document}